\documentclass[letterpaper]{article} 
\usepackage{aaai2026}  
\usepackage{times}  
\usepackage{helvet}  
\usepackage{courier}  
\usepackage[hyphens]{url}  
\usepackage{graphicx} 
\usepackage{natbib}  
\usepackage{caption} 
\usepackage{algorithm}
\usepackage{algorithmic}
\usepackage{multirow}
\usepackage{booktabs}
\usepackage{amsfonts}
\usepackage{amsmath}
\usepackage{subcaption}
\usepackage{newfloat}
\usepackage{listings}
\DeclareCaptionStyle{ruled}{labelfont=normalfont,labelsep=colon,strut=off} 
\floatstyle{ruled}
\newfloat{listing}{tb}{lst}{}
\floatname{listing}{Listing}
\nocopyright

\title{Epi$^2$-Net: Advancing Epidemic Dynamics Forecasting with Physics-Inspired Neural Networks}
\author {
    Rui Sun\textsuperscript{\rm 1} \equalcontrib,
    Chenghua Gong \textsuperscript{\rm 1} \equalcontrib,
    Tianjun Gu\textsuperscript{\rm 2},
    Yuhao Zheng\textsuperscript{\rm 1},
    Jie Ding\textsuperscript{\rm 1},
    Juyuan Zhang\textsuperscript{\rm 1},
    Liming Pan\textsuperscript{\rm 1},
    Linyuan Lü\textsuperscript{\rm 1} \thanks{Linyuan Lü is the corresponding author.}
}
\affiliations {
    \textsuperscript{\rm 1}University of Science and Technology of China\\
    \textsuperscript{\rm 2}East China Normal University\\
    \{rrsun, gongchenghua\}@mail.ustc.edu.cn, \{pan\_liming, linyuan.lv\}@ustc.edu.cn
}

\usepackage{bibentry}

\begin{document}

\maketitle

\begin{abstract}
Advancing epidemic dynamics forecasting is vital for targeted interventions and safeguarding public health.
Current approaches mainly fall into two categories: mechanism-based and data-driven models.
Mechanism-based models are constrained by predefined compartmental structures and oversimplified system assumptions, limiting their ability to model complex real-world dynamics, while data-driven models focus solely on intrinsic data dependencies without physical or epidemiological constraints, risking biased or misleading representations.
Although recent studies have attempted to integrate epidemiological knowledge into neural architectures, most of them fail to reconcile explicit physical priors with neural representations.
To overcome these obstacles, we introduce Epi$^2$-Net, a \textbf{Epi}demic Forecasting Framework built upon \textbf{P}hysics-\textbf{I}nspired Neural \textbf{Net}works.
Specifically, we propose reconceptualizing epidemic transmission from the physical transport perspective, introducing the concept of neural epidemic transport.
Further, we present a physic-inspired deep learning framework, and integrate physical constraints with neural modules to model spatio-temporal patterns of epidemic dynamics.
Experiments on real-world datasets have demonstrated that Epi$^2$-Net outperforms state-of-the-art methods in epidemic forecasting, providing a promising solution for future epidemic containment.
The code is available at: \url{https://anonymous.4open.science/r/Epi-2-Net-48CE}
\end{abstract}

\section{Introduction}

Epidemic transmission poses systemic threats to both human health and societal development.
Prolonged SARS-CoV-2 exposure induces not only acute respiratory syndromes but also cardiovascular complications and immune dysfunction~\cite{xie2022long,al2024long}.
Research confirms the COVID-19 pandemic has caused structural economic damage that may persist for decades, with long-term losses surpassing those of regional conflicts~\cite{jorda2022longer}.
More profound impacts emerge in education, where COVID-19 school disruptions potentially reducing the present value of this cohort's lifetime earnings by 17 trillion dollars globally~\cite{azevedo2021state}.
These multidimensional consequences highlight the critical need for advanced epidemic forecasting models to enable timely health interventions and public policy making. 

Epidemic dynamics modeling is currently guided by two primary methodological paradigms: mechanism-based models and data-driven approaches.
Mechanism-based models are grounded in epidemiological theory and employ systems of differential equations to characterize disease transmission dynamics.
For example, the well-known SIR (Susceptible-Infectious-Recovered) model~\cite{smith2004sir,cooper2020sir} and it variants~\cite{heng2020approximately,poonia2022enhanced} solve partial or ordinary differential equations (PDEs/ODEs) in simulating epidemic transitions between population compartments to generate epidemic predictions.
While providing clear epidemiological interpretability, 
such models incur substantial computational costs when applied at large spatial scales~\cite{bauer2015quiet} and require careful calibration of key parameters to accurately capture real-time epidemic dynamics.
Fundamentally, most mechanism-based models rely on idealized closed-system assumptions~\cite{smith2004sir} and fail to account for the openness and complexity of real-world epidemic spread, including significant factors such as human behavior~\cite{panagopoulos2021transfer,hy2022temporal}, social determinants~\cite{nguyen2023predicting} and geo-spatial interactions~\cite{fan2023exploring}, both of which have been proven to significantly alter transmission trajectories.

In contrast, data-driven methods excel at uncovering latent dependencies within historical surveillance data without explicit formulation of spreading mechanisms~\cite{liu2024review}. 
Conventional statistical methods~\cite{mahmud2020bangladesh,kufel2020arima} provide a foundational framework for data-driven epidemic forecasting through extracting temporal patterns from time series of daily reported cases.
Further, classical machine learning methods have proven effective in capturing the nonlinear temporal dependencies inherent in epidemic dynamics~\cite{wang2020prediction,chimmula2020time,galasso2022random,fang2022application}, leading to more accurate and robust predictions of outbreaks.
With the development of deep learning techniques, recent breakthroughs in neural architectures are revolutionizing data-driven epidemic forecasting.
By automatically extracting spatio-temporal signatures from diverse data streams (case reports, mobility traces, and geographic metadata), deep learning methods offer transformative potential in modeling epidemic dynamics in real-world scenarios ~\cite{panagopoulos2021transfer,fan2023exploring,liu2024review}.
However, the absence of physical constraints and epidemiological mechanisms in data-driven models may introduce incorrect inductive biases in data-sparse situations, such as during epidemic variations and early outbreaks~\cite{punn2020covid}.
Moreover, data-driven models also suffer from intrinsic interpretability limitations, challenging their adoption for public health policy-making and government intervention measures~\cite{gong2025epillm,zhou2025ph}.

Consequently, the integration of mechanistic principles with data-driven approaches, particularly through physics-inspired deep learning~\cite{raissi2019physics}, represents a key direction for advanced epidemic forecasting.
Although recent studies have attempted to integrate epidemiological knowledge or physical constraints into neural
architectures, most of them fail to reconcile explicit physical priors with neural representations~\cite{chen2018neural} and exhibit critical limitations as follows:

\textbf{1) Discrepancy between physical constraints and real-world epidemics.} 
Real-world epidemic dynamics involve complex factors such as human mobility, geographical distance, and lockdown interventions.
Currently, mainstream physics-inspired neural methods couple SIR equations with deep learning modules, thereby endowing neural architectures with  epidemiological mechanistic~\cite{kosma2023neural,rodriguez2023einns}. 
Despite providing certain physical constraints, the SIR model's inherent closed-system assumption leads to unrealistic simplifications for real-world epidemics~\cite{tian2024air}.
Such models fail to capture cross-border transmission and spatial attenuation driven by human mobility or geographic factors~\cite{wanearth2025}.
Furthermore, the rigid assumptions of closed systems exhibit time-varying limitations, failing to represent uncertainties including intervention effects, viral mutations, and temporal fluctuations in real-world epidemic transmission~\cite{du2024advancing}.
Given this mismatch, how to propose more generalizable physical priors to guide the design of neural architecture becomes a central challenge.

\textbf{2) Discrepancy between physical constraints and neural representations.} 
The success of physics-inspired neural networks lies in reconciliation of explicit physical priors with neural representations~\cite{raissi2019physics,chen2018neural}.
For epidemic forecasting, current physics-inspired deep learning methods typically use neural modules to encode the spatio-temporal epidemic patterns and forcibly align the obtained representations with explicit physical quantities~\cite{gao2021stan,kosma2023neural}. 
Nevertheless, the latent representation space typically involves nonlinear transformations and combinations, leading to inherent discrepancy when mapped to specific physical variables~\cite{tian2024air}.
To ensure the model adheres to physical principles while retaining data-driven flexibility, the key architectural design aims to align implicit neural representations with well-established physical meanings~\cite{choi2023climate,verma2024climode,hettige2024airphynet}, thereby bridging this discrepancy. 

In this paper, we present a novel physics-inspired deep learning framework for epidemic dynamics modeling named Epi$^2$-Net, which integrates physical principles of epidemic into neural networks, harmonizing the inconsistency between mechanism-based and data-driven methods.
Notably, we reconceptualize epidemic transmission through the lens of physics transport, and establish connections between epidemic modeling and the well-studied diffusion-advection-reaction (DAR) phenomenon~\cite{cosner2014reaction}, which have been rigorously applied in meteorological~\cite{tian2024air,du2025graph} and marine science~\cite{adam2004use}.
Through systematic mapping of physical transport mechanisms to epidemic transmission patterns, we introduce the neural epidemic transport equation to generalize the complexity of real-world epidemic scenarios.
Building on the neural epidemic transport equation, we further incorporate a Neural ODE module and neural networks to model spatiotemporal dependencies of epidemics. This design inherently combines physical consistency with adaptability and flexibility of neural representations.
Our contributions are summarized as follows:

\begin{itemize}
    \item We provide a novel view of physical transport for epidemic dynamics modeling and introduce the Neural Epidemic Transport equation to generalize the complexity of real-world epidemic scenarios.
    
    \item 
    We propose Epi$^2$-Net, a physics-inspired deep learning framework that synergistically integrates physics priors with neural architectures, unifying interpretable physics guidance with data-driven representational flexibility.

    \item We conduct experiments on four real-world COVID-19 datasets and show that Epi$^2$-Net outperforms the state-of-the-art competitors in epidemic forecasting task.
\end{itemize}

\section{Related Work}

\subsection{Epidemic Dynamics Modeling}

\paragraph{Mechanism-based models.}
Mechanism-based models integrate physical priors with differential equations
to characterize epidemics under idealized conditions, with SIR and its variants being prime examples~\cite{cooper2020sir}.
While offering interpretability, they suffer from fundamental limitations of closed-system assumptions~\cite{smith2004sir}.
The oversimplified physical system demonstrates insufficient generalizability when facing complex, real-world epidemic scenarios.
Specifically, the idealized contact assumption neglects natural population mobility~\cite{panagopoulos2021transfer,hy2022temporal} and geographical dependencies~\cite{fan2023exploring}, thereby failing to capture urban diffusion dynamics and human activity-driven transmission.
Furthermore, factors like viral mutations, vaccine uptake, and policy interventions pose challenges for the real-time calibration of mechanistic methods~\cite{du2024advancing}. 
Although researchers have attempted to address these issues through mechanistic refinements~\cite{heng2020approximately,poonia2022enhanced}, these adjustments remain inadequate for modeling intricate spatiotemporal patterns and may introduce incorrect inductive biases~\cite{wu2018deep}.

\paragraph{Data-driven models.}
Given the intricate interplay of real-world pandemic factors, data-driven models identify latent statistical patterns from observational data to forecast epidemic dynamics.
Statistical models such as ARIMA~\cite{kufel2020arima} and PROPHET~\cite{mahmud2020bangladesh} predict future epidemic trends by analyzing temporal patterns of case number; machine learning models including Linear Regression~\cite{kaur2022forecasting}, Random Forest~\cite{galasso2022random}, and XGBOOST~\cite{fang2022application} enhance forecasting by modeling complex nonlinear relationships in historical observational records; while deep learning techniques (e.g., LSTM, Transformer, and GNN) further capture spatio-temporal transmission patterns from multivariate data sources~\cite{chimmula2020time,liu2024review}, such as social
determinants~\cite{nguyen2023predicting}, geographical interactions~\cite{fan2023exploring}, and regional mobility~\cite{panagopoulos2021transfer}.
However, purely data-driven approaches risk violating essential epidemiological principles, potentially generating biologically implausible dynamics.

\subsection{Physics-Inspired Deep Learning}
Recent advances prioritize physics-inspired neural networks to combine domain knowledge with powerful deep learning models to solve complex scientific and engineering problems~\cite{karniadakis2021physics,karpatne2024knowledge}.
One implementation constrains the loss function with physical priors via penalty terms, ensuring outputs comply with this knowledge but often fragmenting physical constraints from representation learning~\cite{shi2021physics}.
More advanced alternatives involve designing hybrid architectures that integrate knowledge into specific model components, thereby bridging the discrepancy between explicit physical equations and implicit neural representations~\cite{verma2024climode,hettige2024airphynet,nascimento2021hybrid}.
Notably, the continuity equation, renowned for its modeling capacity and extensibility for real-world scenarios, has been widely adopted in this manner across atmospheric sciences~\cite{hettige2024airphynet}, marine science~\cite{adam2004use} and transportation engineering~\cite{ji2022stden}.
For epidemic dynamics modeling, we pioneer the adoption of physical transport processes, derived from the continuity equation, to guide physics-inspired model design, given the aforementioned two discrepancies for physic-inspired neural methods.

\section{Preliminaries}
\subsection{Problem Definition}
In this paper, we formulate epidemic dynamics forecasting as a spatio-temporal prediction task.
Formally, 
let $X_{1:T} = \{ X_t \}_{t=1}^T \in \mathbb{R}^{N \times D \times T}$ denote historical epidemiological observations, where $N$ is the number of regions, $T$ is the historical time steps, $D$ is the feature dimension ($D =1$ for daily confirmed cases in our study), and $x_t \in \mathbb{R}^{N \times D}$ is the multivariate regional observations at time $t$.
Let $G = (V,E)$ denote the inter-region graph, where $V$ is the set of regions, $E$ is the set of edges, $A$ denotes the geospatial adjacency matrix of $G$ and $A[i,j]$ denotes the spatial distance between regions $i$ and $j$.
Moreover, human mobility patterns of regions are defined as $M_{1:T}=\{ M_t \}_{t=1}^T \in \mathbb{R}^{N \times N \times T}$, where $M_t[i,j]$ quantifies the mobility from region $i$ to $j$ at time $t$. 
The objective is to learn a function $f(\cdot)$ that can accurately predict future epidemic trends over horizon $h$:
\begin{equation}
    \hat{X}_{T+1:T+h} = f(X_{1:T},M_{1:T},A).
\end{equation}

\subsection{Concepts of Physical Transport}
This paper pioneers a physical transport perspective for epidemic modeling. First, we begin by introducing the foundational concepts of physical transport.

\paragraph{Continuity Equation.} 
The continuity equation expresses the conservation of a quantity in a physical system.
Specifically, it quantifies the relationship between the temporal variation of system mass and the inflow/outflow of matter:
\begin{equation}
    \frac{\partial c}{\partial t} + \nabla \cdot (c\vec{v})  = 0,
\end{equation}
where $c$ is the fluid density, $\vec{v}$ is the velocity vector, and $\nabla \cdot$ is the divergence operator.
Based on this equation, extensive physical transport processes like diffusion, advection, and reaction can be derived.

\paragraph{Diffusion.}
Diffusion describes the process where particles randomly move from high to low concentration areas in a medium, which can be formulated by continuity equation and Fick's Law~\cite{grady2010discrete,paul2014thermodynamics}:
\begin{equation}
    \frac{\partial c}{\partial t} \bigg|_{dif} -  k \nabla^2 c = 0,
\end{equation}
where $k$ is the diffusion coefficient, and $\nabla^2$ is a Laplacian operator combining divergence and gradient.

\paragraph{Advection.} 
Distinct from the diffusion, advection characterizes the bulk transport of particles driven by an imposed velocity field in fluid systems~\cite{grady2010discrete}:
\begin{equation}
    \frac{\partial c}{\partial t} \bigg|_{adv} + \vec{v} \cdot \nabla c = 0,
\end{equation}
where $\vec{v}$ denotes the velocity vector field. 

\paragraph{Reaction.} 
Reaction characterizes the localized substance transformations that alter concentrations~\cite{volpert2009reaction}, governed by:
\begin{equation}
    \frac{\partial c}{\partial t} \bigg|_{rea} - R(c,t) = 0,
\end{equation}
where the reaction term $R(\cdot)$ quantifies the source/sink dynamics from physicochemical or other transformations.

\paragraph{Diffusion-Advection-Reaction.}
Through coupling of diffusion, advection and reaction, we complete the Diffusion-Advection-Reaction (DAR) triad~\cite{cosner2014reaction}: 
\begin{equation}
    \frac{\partial c}{\partial t} \bigg|_{DAR} = \underbrace{k \nabla^2 c}_{diffusion} - \underbrace{\vec{v} \cdot \nabla c}_{advection} + \underbrace{R(c,t)}_{reaction}.
\end{equation}
Widely applied in various fields, DAR systems in prior works rely on low-dimensional, handcrafted, and non-neural systems, determined by domain experts through trial and error~\cite{adam2004use,khokhlov1995propagation}.
\begin{figure}[ht]
    \centering 
        \centering
\includegraphics[width=0.43\textwidth]{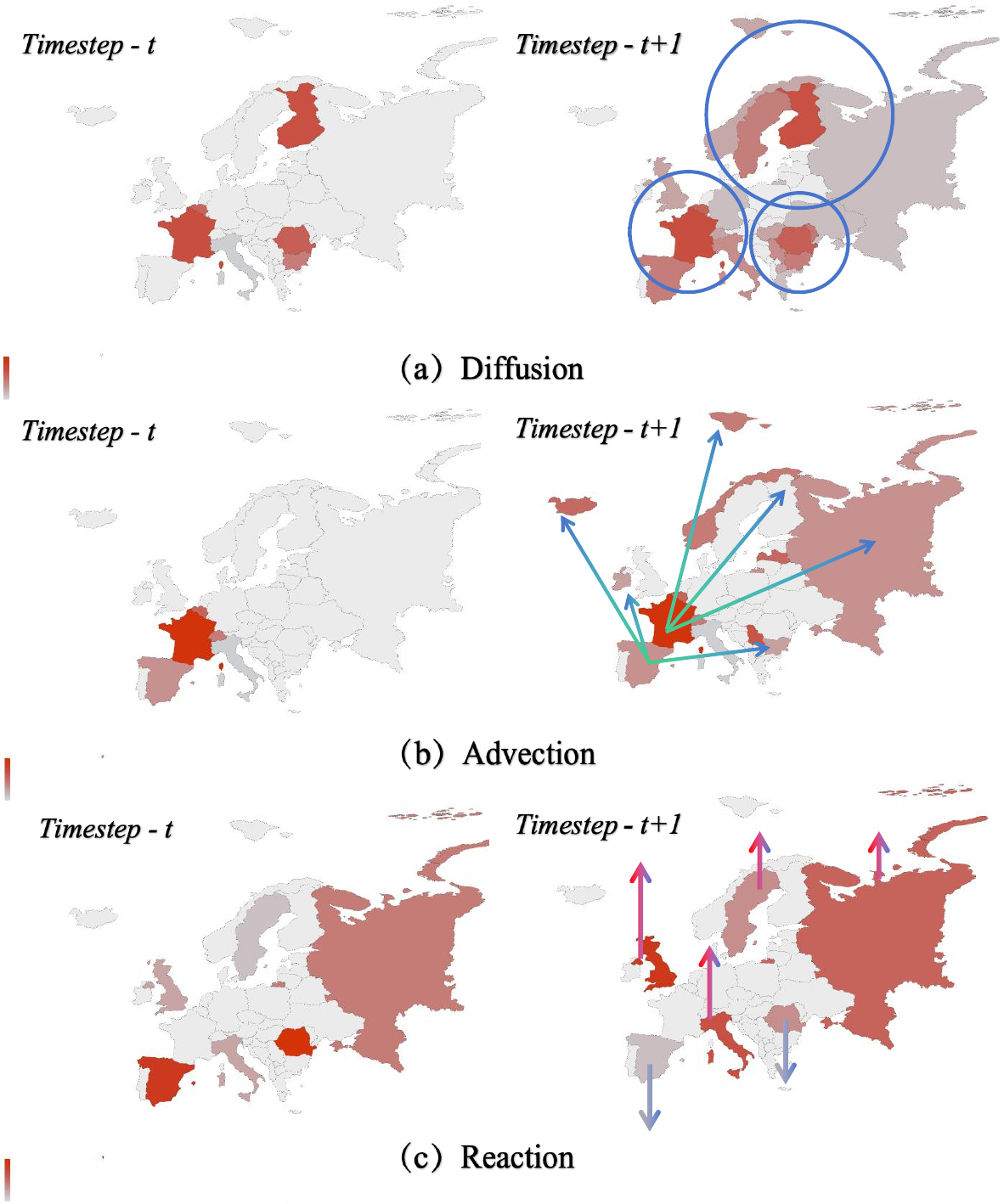} 
    \caption{Motivation of introducing physical transport into epidemic dynamics modeling. The heatmaps show infection density for regions at timestep $t$ and $t+1$. Diffusion illustrates the  spread from severely affected regions to neighboring regions; advection captures the spread to specific regions due to human mobility; reaction signifies localized shifts in the regional epidemic, influenced by interventions like lockdown policies, viral mutations, and vaccination.} 
    \label{mov} 
\end{figure}
\begin{figure*}[ht]
    \centering 
        \centering
        \includegraphics[width=0.73\textwidth]{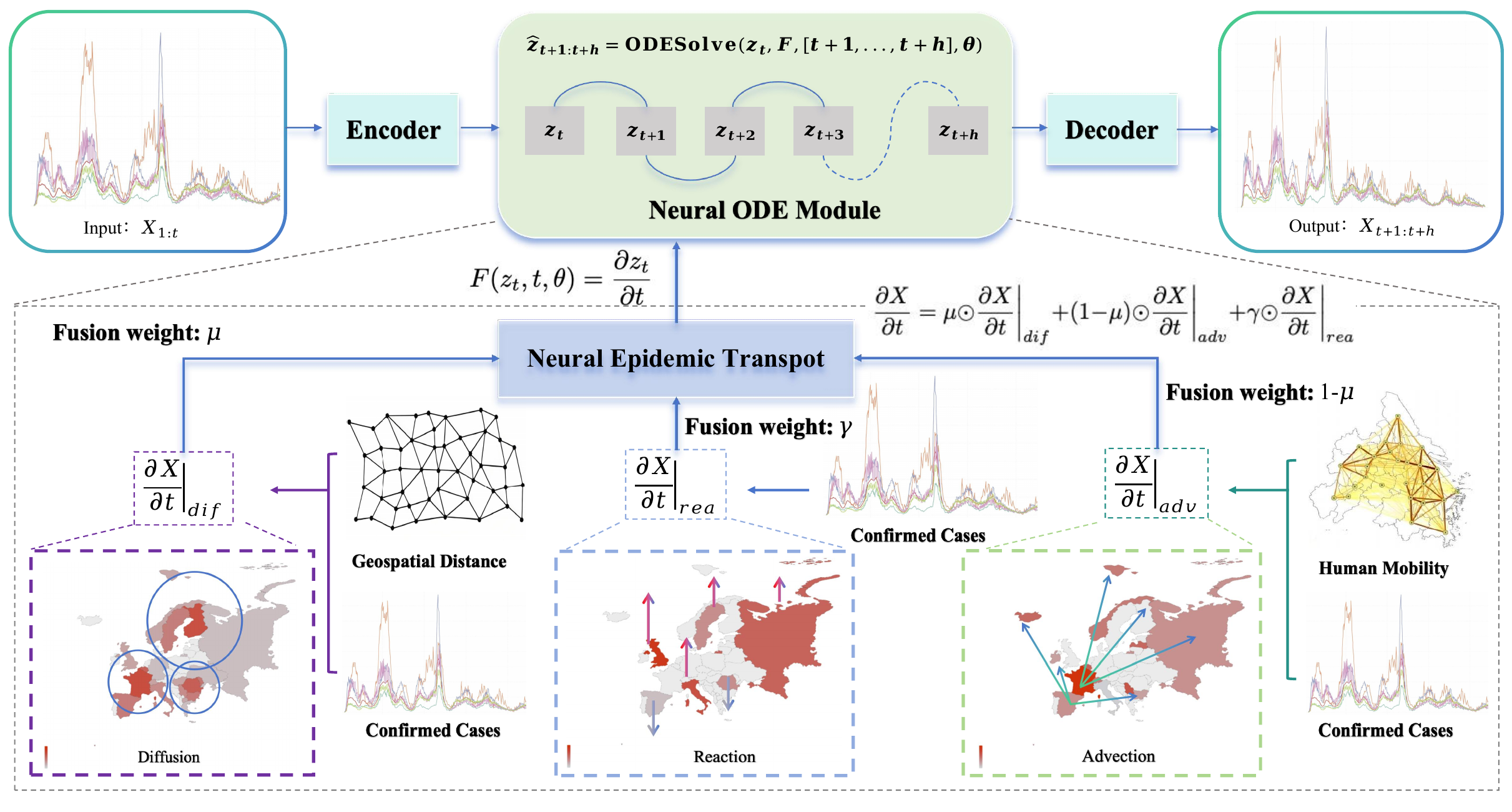} 
    \caption{The overall framework of Epi$^2$-Net consists of three components: 1) an encoder to model temporal dependency from the historical observations, 2) a Neural ODE module to module the evolution of epidemic dynamics, 3) a decoder to generate predictions of epidemic trends based on the neural representations and physical priors.} 
    \label{framework} 
\end{figure*}

\subsection{Neural Epidemic Transport}
Inspired by recent studies~\cite{verma2024climode,hettige2024airphynet}, we view epidemic dynamics as a flux, a spatial movement of quantities over time, for further modeling. 
To accurately model complex real-world epidemics, we integrate three physical transport processes with the intricacies of epidemic transmission, thereby defining the neural epidemic transport equation.
A more intuitive understanding of our motivation is depicted in Figure \ref{mov}.

\paragraph{Diffusion.}
The diffusion of epidemics describes the spread from regions with high infection density to neighboring regions with low density, analogous to heat conduction or gas diffusion, with a focus on characterizing the spatial dependence of epidemic transmission.
Following~\cite{bronstein2017geometric}, we derive the discrete diffusion equation in terms of the graph Laplacian operator:
\begin{equation}
    \frac{\partial X^i}{\partial t}\bigg|_{dif} = k \cdot \sum_{j\in N(i)}G^{dif}[i,j](X^i-X^j)  = k \cdot L X,
\end{equation}
where $X^i$ denotes the epidemic density of region $i$, $N(i)$ denotes the neighboring set of $i$, and $G^{dif}[i,j]$ denotes the transport weight of diffusion from $i$ to $j$.
In our implementation, we compute the spatial distances between regions based on centroid coordinates to establish the diffusion graph $G^{dif}[i,j] = \frac{1}{d_{ij}}$, where $d_{ij}$ is the haversine distance between regions $i$ and $j$.
$L$ denotes the normalized graph Laplacian matrix, derived from the static $G^{dif}$(equivalent to the geospatial adjacency matrix $A$).

\paragraph{Advection.} 
The advection of epidemics refers to the rapid, cross-regional transmission of pathogens driven by large-scale, directional population movements (e.g., transportation, migration), analogous to the pollutant transport driven by river currents or atmospheric flow, with particular emphasis on characterizing the spatio-temporal dynamics of pandemic transmission. 
Following~\cite{chapman2015advection}, the discrete analogue of advection can be formulated as:
\begin{equation}
    \frac{\partial X^i}{\partial t}\bigg|_{adv} = \sum_{\forall j | j\rightarrow i} G^{adv}[j,i] X^j  - \sum_{\forall i | i \leftarrow k} G^{adv}[i,k] X^i,
\end{equation}
where $G^{adv}[i,j]$ denotes the transport weight of advection from region $i$ to $j$.
As population movement as a critical factor in epidemic spread, we model the human mobility as the flow field in advection, following~\cite{hettige2024airphynet}. This allows us to construct the advection graph using inter-regional human flow data:
\begin{equation}
    \frac{\partial X}{\partial t}\bigg|_{adv} = M_tX,
\end{equation}
where $M_t$ quantifies the cross-regional mobility at time $t$, and $M$ includes dynamic and directed graph snapshots.

\paragraph{Reaction.}
The reaction of epidemics indicates the complex intra-regional dynamics, such as localized lockdowns, vaccination campaigns, herd immunity, and viral variants, which depend on both biological factors and social determinants in real-world scenarios.
Distinct from diffusion and advection, the reaction primarily models endogenous patterns within a region, which can be analogized to autocatalytic and equilibrium inhibition phenomena of chemical reactions~\cite{bissette2013mechanisms,lewis1925new}:

\begin{equation}
    \frac{\partial X}{\partial t}\bigg|_{rea} = R(X),
\end{equation}
where $R(\cdot)$ denotes the reaction term. Given the flexibility of localized dynamics, modeling it using learnable parameters through neural network is a viable approach~\cite{du2025graph}. 

\paragraph{Neural epidemic transport equation.}
Based on the previous discussion, we introduce the neural epidemic transport equation, which unifies real-word epidemic complexity including geographic dependencies, human mobility, and localized reaction dynamics:
\begin{equation}
    \frac{\partial X}{\partial t} =  \mu \odot \frac{\partial X}{\partial t} \bigg|_{dif} + (1-\mu) \odot \frac{\partial X}{\partial t} \bigg|_{adv} + \gamma \odot \frac{\partial X}{\partial t} \bigg|_{rea},
\end{equation}
where $\mu$ and $\gamma$ are learnable coefficients obtained by neural networks.
Given the intricate interplay between diffusion and advection in particle transport~\cite{hettige2024airphynet}, we incorporate a learnable parameter $\mu$ for adaptive fusion to highlight inter-regional patterns. 
Concurrently, under an open-system assumption~\cite{tian2024air}, the reaction parameter $\gamma$ prioritizes the intra-regional epidemic modeling.

\subsection{Physics-Informed Neural Framework}
Based on the neural epidemic transport, we further introduce the physics-inspired neural framework, Epi$^2$-Net, which includes three main components: a encoder, a Neural ODE module, and a decoder.
The overall architecture of Epi$^2$-Net is depicted in Figure \ref{framework}.

\paragraph{Encoder.}
To fully leverage the data-driven flexibility, we employ an neural encoder to extract temporal dependencies from the historical epidemic observations, and obtain the initial state for neural ODE module~\cite{chen2018neural}:
\begin{equation}
    z_{t} = \text{Encoder}(X_{t-w:t}),
\end{equation}
where $w$ denotes the observation window length. 
Following prior studies~\cite{ji2022stden,hettige2024airphynet}, we implement the encoder by a RNN-based module combined with a reparameterization trick~\cite{kingma2013auto}.

\paragraph{Neural ODE Module.}
To integrate physics constraints into neural framework, we 
draw inspiration from Neural ODEs~\cite{chen2018neural} and introduce the neural epidemic transport to guide the module design.
Specifically, our Neural ODE vector field $F(\cdot)$ is formulated using the discretized graph operators and reaction term, as follows:
\begin{equation}
    F(z_t,t,\theta) = \frac{\partial z_{t}}{\partial t} = \mu \odot (-k\cdot L X ) - (1-\mu) \odot (MX) + \gamma \odot X, 
\end{equation}
where $\odot$ is the hadamard product operation, $\theta$ denotes learnable parameters to approximate the neural epidemic transport.
The diffusion coefficient $k$ and reaction coefficient $\gamma$ are updated by neural networks.
Moreover, the fusion weight $u$ is obtained by a fully-connected layer, leveraging neural representations of diffusion and advection branches.
After reconciling physical priors with neural representations, we apply the Neural ODE solver to compute the future states:
\begin{equation}
    \hat{z}_{t+1: t+h} = \text{ODESolve}(z_t,F,[t+1,...,t+h],\theta).
\end{equation}

\paragraph{Decoder.}
Finally, a decoder generates epidemic predictions from the latent representation.
Consistent with~\cite{hettige2024airphynet,ji2022stden}, we utilize the non-neural decoding operation (reshaping and aggregation) to produce the output format at each time step:
\begin{equation}
    \hat{X}_{t+1;t+h} = \text{Decoder}(\hat{z}_{t+1: t+h}).
\end{equation} 
Subsequently, we optimizes Epi$^2$-Net via back-propagation to minimize the Root Mean Square Error (RMSE) function:
\begin{equation}
    \mathcal{L} = \sqrt{\frac{1}{h} \sum^{t+h}_{k=t+1}  \Vert X_k - \hat{X}_k\Vert_2^2} .
\end{equation}

\section{Experiments}

\subsection{Experimental Settings}
\paragraph{Datasets.}
We evaluate our proposed Epi$^2$-Net on four real-world COVID-19 datasets: England, France, Italy, and Spain~\cite{panagopoulos2021transfer}.
The dataset comprises three main components: daily cases, human mobility, and geographical distances. 
We obtained daily cases and human mobility data from open-source GitHub repository\footnote{https://github.com/geopanag/pandemic\_tgnn}, and manually constructed geographical distance information.
According to~\cite{hy2022temporal,nguyen2023predicting,gong2025epillm}, we employ the daily reported case as the epidemic feature.
The target is to forecast the new cases in specific regions for given countries.
Basic statistics of dataset and further details are provided in Appendix A.

\paragraph{Settings and implementations.}
To comprehensively evaluate Epi$^2$-Net, we conduct
epidemic predictions across three horizons: 3 days, 5 days, and 7 days.
We split each dataset chronologically in the ratio of 60\%:20\%:20\% to generate training, validation, and test sets respectively.
To evaluate performance, we use Mean Absolute Error (MAE) and Root Mean Square
Error (RMSE) as metrics (see in Appendix B).
More implementation details are provided in Appendix C.

\paragraph{Baselines.}
We evaluate Epi$^2$-Net against 14 representative baselines from four categories:
1) \textbf{Statistical methods}: 
HA (Historical Average), PROPHET~\cite{mahmud2020bangladesh}, and ARIMA~\cite{kufel2020arima}. 
2) \textbf{Machine learning methods}: LIN\_REG~\cite{kaur2022forecasting}, RAND\_FOREST~\cite{galasso2022random} XGBOOST~\cite{fang2022application}.
3) \textbf{Deep learning methods}: ITRANSFORMER~\cite{liu2023itransformer}, TIMEMIXER~\cite{wang2024timemixer}, 
STGNN~\cite{kapoor2020examining}, ATMGNN~\cite{nguyen2023predicting}
4) \textbf{Physics-inspired neural methods}:
ODE-LSTM~\cite{lechner2020learning},  
GN-ODE~\cite{kosma2023neural},
EINNs~\cite{rodriguez2023einns},
EARTH~\cite{wanearth2025}.
More baseline details are provided in Appendix D.
\begin{table*}[ht]
    \caption{The overall performance comparison of epidemic dynamics forecasting on metric MAE. The bold and underlined fonts show the best and the second best result. The improvement(\%) is calculated based on these two results.}
    \label{results-mae}
    \small
    \setlength{\tabcolsep}{2mm}
    \renewcommand{\arraystretch}{0.9}
    \begin{tabular}{c | c c c | c c c | c c c | c c c }
    \toprule
    \multirow{2}{*}{\textbf{Model}} & 
    \multicolumn{3}{c|}{\textbf{England}} &
    \multicolumn{3}{c|}{\textbf{France}} &
    \multicolumn{3}{c|}{\textbf{Italy}} &
    \multicolumn{3}{c}{\textbf{Spain}} \\ 
    \cmidrule(lr){2-4} \cmidrule(lr){5-7} \cmidrule(lr){8-10} \cmidrule(lr){11-13}
    & 3 days & 5 days & 7 days & 3 days & 5 days & 7 days & 3 days & 5 days & 7 days & 3 days & 5 days & 7 days \\
    \midrule
        HA & 7.24 & 7.63 & 7.97 & 6.03 & 6.08 & 6.21 & 30.89 & 30.69 & 29.59 & 74.03 & 74.35 & 74.27 \\
        PROPHET & 10.19 & 10.52 & 14.39 & 8.07 & 8.17 & 8.41 & 46.79 & 50.05 & 51.29 & 85.59 & 86.99 & 86.57 \\
        ARIMA & 8.13 & 8.08 & 8.72 & 6.02 & 6.24 & 6.34 & 45.81 & 46.29 & 47.52 & 84.32 & 85.53 & 85.01 \\
        \midrule
        LIN\_REG & 8.40 & 9.10 & 10.11 & 6.74 & 6.68 & 6.20 & 31.94 & 32.50 & 33.29 & 72.39 & 73.45 & 78.05 \\
        RAND\_FOREST & 7.52 & 7.18 & 7.00 & 4.17 & 4.05 & 4.53 & 29.37 & 29.42 & 29.17 & 53.32 & 55.81 & 53.42 \\
        XGBOOST & 7.40 & 6.76 & 7.09 & 4.56 & 4.97 & 4.86 & 27.75 & 28.18 & 28.60 & 50.07 & 50.74 & 52.55 \\
        \midrule
        ITRANSFORMER & 6.12 & 6.26 & 6.45 & 2.93 & 3.04 & 3.13 & 21.09 & 21.52 & 21.58 & 29.91 & 30.53 & 30.95 \\
        TIMEMIXER & \underline{5.95} & 6.16 & 6.31 & \underline{2.86} & \underline{3.07} & 3.29 & 21.08 & 21.46 & 21.80 & 28.85 & 28.13 & 28.44 \\
        STGNN & 6.70 & 6.77 & 6.56 & 3.66 & 3.20 & 3.35 & 25.89 & 26.72 & 26.85 & 27.55 & 27.33 & 27.06 \\
        ATMGNN & 6.08 & 6.21 & \underline{6.08} & 3.44 & 3.34 & \underline{3.09} & 24.63 & 24.68 & 24.02 & \underline{26.05} & \underline{26.52} & \underline{26.74} \\
    \midrule
        ODE-LSTM & 8.15 & 8.39 & 8.66 & 5.04 & 5.21 & 5.84 & 30.82 & 30.77 & 31.05 & 41.94 & 42.00 & 42.08 \\
        GN-ODE & 8.25 & 8.48 & 9.54 & 5.92 & 5.87 & 5.91 & 27.57 & 27.01 & 27.90 & 42.63 & 41.99 & 47.70 \\
        EPINNs & 7.23 & 8.63 & 8.72 & 3.02 & 3.99 & 3.92 & 21.02 & 2.32 & 21.57 & 28.95 & 28.24 & 28.99 \\
    EARTH & 5.96 & \underline{6.05} & 6.22 & 3.96 & 3.78 & 3.49 & \underline{20.05} & \underline{20.37} & \underline{20.74} & 26.61 & 26.75 & 26.93 \\
    Epi$^2$-Net & \textbf{5.35} & \textbf{5.50} & \textbf{5.70} & \textbf{2.55} & \textbf{2.62} & \textbf{2.44} & \textbf{18.14} & \textbf{18.21} & \textbf{18.53} & \textbf{25.59} & \textbf{25.19} & \textbf{24.18} \\
    \midrule
    Improvement(\%) & 10.08 & 9.09 & 6.25 & 10.84 & 14.65 & 21.03 & 9.52 & 10.60 & 10.65 & 1.76 & 5.01 & 9.57 \\
    \bottomrule
    \end{tabular}  
\end{table*}

\subsection{Performance Comparison}
From Table~\ref{results-mae}, we can observe that: 
1) Epi$^2$-Net consistently outperforms all baselines across every dataset, achieving a remarkable 21.03\% improvement in MAE over the best-performing baseline on France-7days.
2) deep learning methods, particularly TIMEMIXER and ATMGNN, outperform statistical and machine learning methods, demonstrate powerful modeling capabilities in capturing complex real-world epidemics.
3) physic-inspired neural methods such as EPINNs and EARTH exhibit competitive performance, indicating benefits of integrating physic priors into neural architectures.
4) the suboptimal performance of physics-inspired neural methods such as ODE-LSTM and GN-ODE stems from oversimplified physical priors, which cannot adapt to multifaceted factors in real-world settings.
Additional experiments on metric RMSE are provided in Appendix E.
\subsection{Case Study}
\begin{figure}[ht]
    \centering 
        \includegraphics[width=0.45\textwidth]{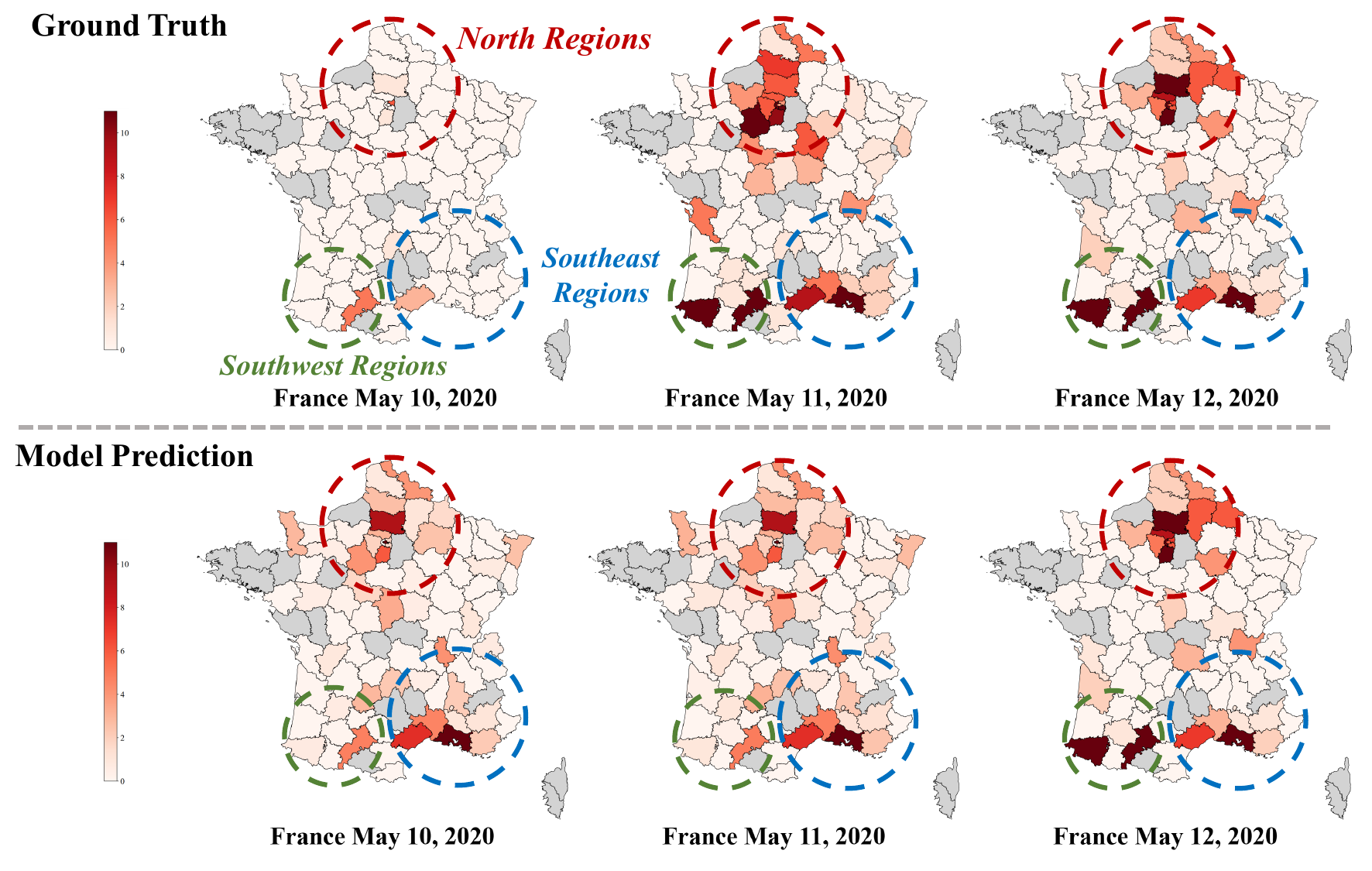} 
    \caption{A case study examines COVID-19 progression in partial regions of France. Gray shading denotes areas lacking surveillance records.} 
    \label{cs_france} 
\end{figure}

\begin{figure}[ht]
    \centering 
        \includegraphics[width=0.45\textwidth]{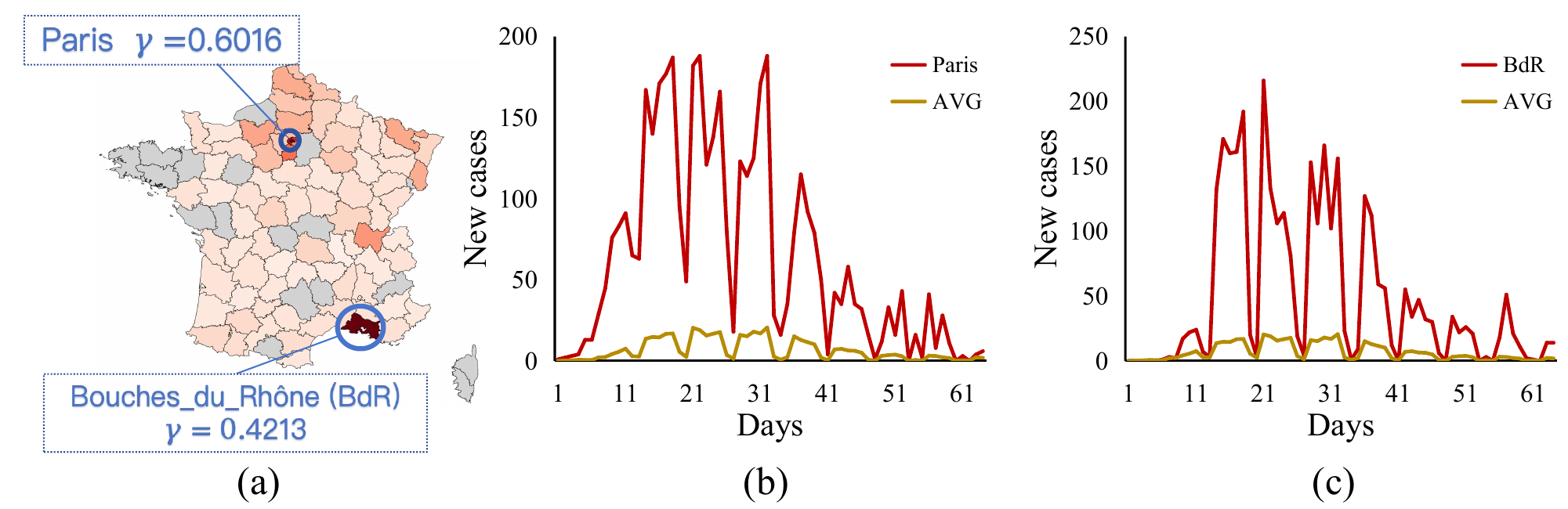} 
    \caption{Visualization of $\gamma$ values and the infection status of two critical regions: Paris and Bouches-du-Rhône (BdR).} 
    \label{cs_gamma} 
\end{figure}

The case study in Figure \ref{cs_france} visualizes the epidemic dynamics of COVID-19 in France. 
We primarily focuses on three key areas: north, southwest, and southeast regions. 
In the early stages of the outbreak (May 10th), EpiNet's predictions already provided early warnings for three major areas. 
By the full escalation of the epidemic (May 12th), EpiNet accurately predicted severely affected areas, supporting precise interventions.
We also visualized the learned $\gamma$ values in Figure \ref{cs_gamma}(a) for each region in France, and performed a fine-grained analysis of Paris and Bouches-du-Rhône (BdR), two regions with the largest $\gamma$. 
Figures \ref{cs_gamma}(b) and \ref{cs_gamma}(c) show the new cases for Paris and BdR, indicating that the infection levels in these two regions far exceed the average for all of France, and the interpretability of reaction coefficients.
For more case study and analysis, please refer to Appendix F.

\subsection{Ablation Study}
\begin{figure}[htbp]
    \centering 
    \begin{subfigure}[b]{0.37\textwidth} 
        \centering
        \includegraphics[width=\textwidth]{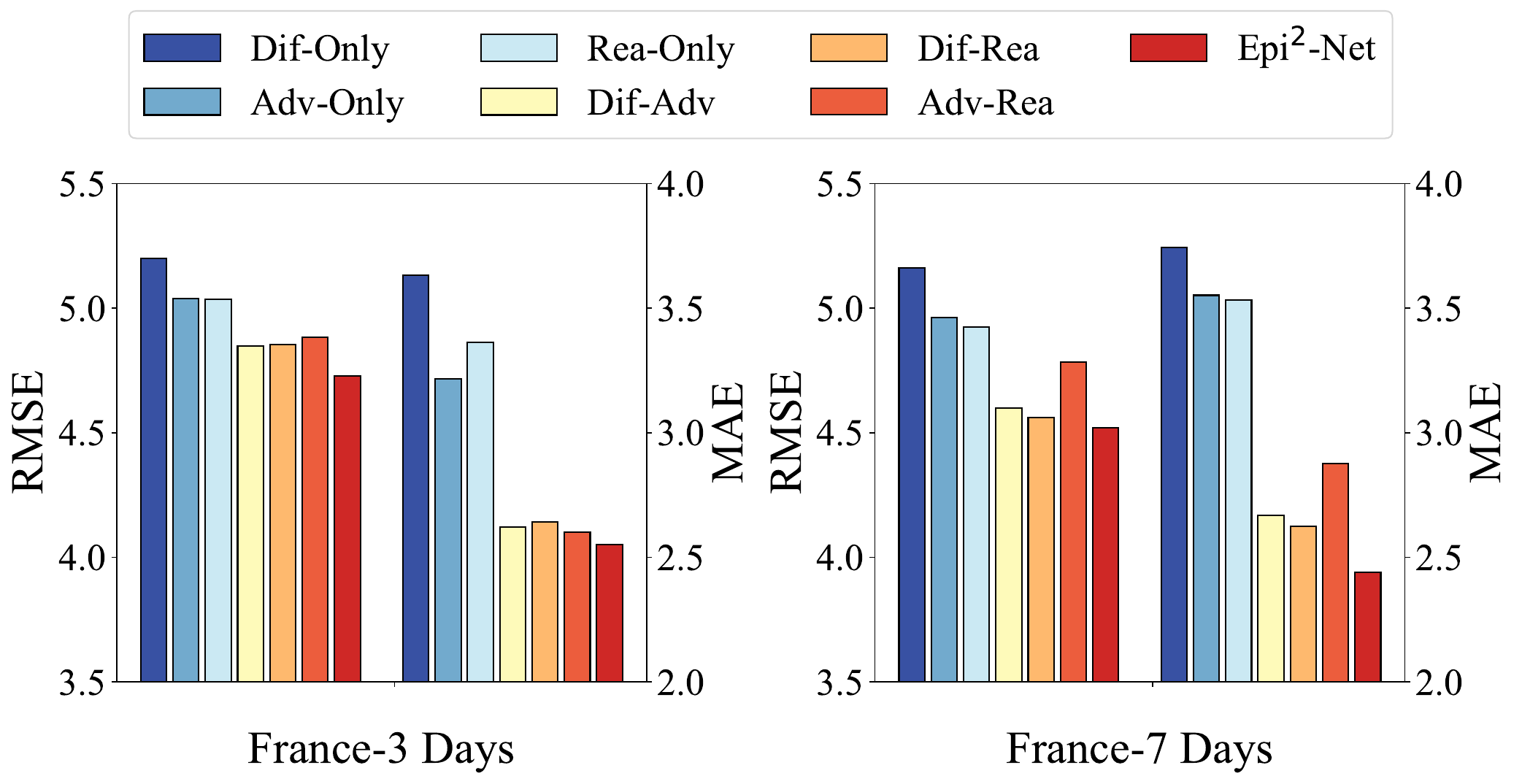} %
    \end{subfigure}
    \hfill 
    \begin{subfigure}[b]{0.37\textwidth}
        \centering
        \includegraphics[width=\textwidth]{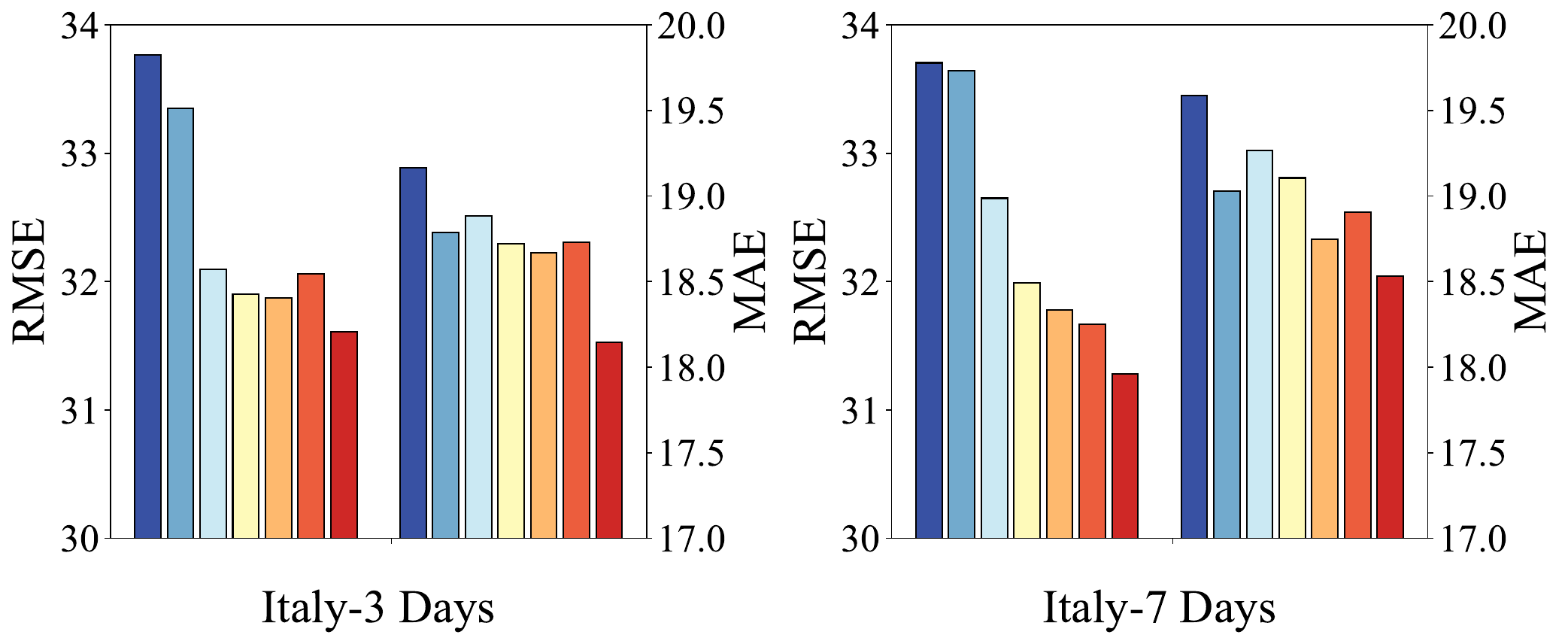}
    \end{subfigure}
    \caption{Ablation study on physical transport processes.} 
    \label{ab1} 
\end{figure}

\paragraph{Physical transport processes.}
To quantify the contribution of physical priors, we compare Epi$^2$-Net against six variants:
1) Dif-Only: only the diffusion process.
2) Adv-Only: only the advection process.
3) Rea-Only: only the reaction process.
4) Dif-Adv: the diffusion and advection processes.
5) Dif-Rea: the diffusion and reaction processes.
6) Adv-Rea: the advection and reaction processes.
Figure \ref{ab1} shows the comparison results on France and Italy, we observe that Epi$^2$-Net achieve the best performance, indicating the significance of integrating comprehensive physic priors (diffusion, advection, and reaction) into epidemic modeling.
Evidently, using single transport process (Dif-Only, Adv-Only, and Rea-Only) as the prior significantly degrades the performance.
While integrating two transport processes (Dif-Adv, Dif-Rea, and Adv-Rea) significantly improves the forecasting accuracy compared to single-process models, the performance remains inferior to Epi$^2$-Net, reaffirming the importance of the complete neural epidemic transport.

\paragraph{Temporal encoder.}
To study the efficacy of the encoder, we compare our model with three variants:
1) LSTM: adopt the LSTM~\cite{hochreiter1997long} as the encoder.
2) TCN: adopt the Temporal Convolutional Network~\cite{bai2018empirical} as encoder.
3) Transformer: adopt the Transformer~\cite{vaswani2017attention} block as encoder.
4) GRU: adopt the Gated Recurrent Unit~\cite{chung2014empirical} as encoder.
Based on Table~\ref{encoder}, GRU temporal encoders are advantageous, as advanced architecture Transformer lose their long-range dependency modeling benefits in this scenario.
For additional experiments of encoder and decoder, please refer to Appendix G.

\begin{table}[htbp]
    \centering
    \caption{Performance comparison of temporal encoder.}
    \small
    \setlength{\tabcolsep}{5pt}
    \renewcommand{\arraystretch}{0.85}
    \begin{tabular}{@{}l cc cc cc @{}} 
        \toprule
        \multirow{2}{*}{Encoder} & \multicolumn{2}{c}{England-3days} & \multicolumn{2}{c}{Spain-5days} & \multicolumn{2}{c}{Italy-7days} \\
        \cmidrule(lr){2-3} \cmidrule(lr){4-5} \cmidrule(lr){6-7}
        &RMSE & MAE & RMSE & MAE & RMSE & MAE \\
        \midrule
        LSTM   & 7.68 & 5.46 & 47.40 & 36.81 & 34.79 & 21.54 \\
        TCN   & 8.58  & 6.56 & 54.70 & 38.89 & 33.30 & 21.23 \\
        Transformer     & 9.32 & 7.63 & 41.77 & 34.02 & 33.62 & 25.18\\
        GRU  & \textbf{7.57} & \textbf{5.35} & \textbf{40.77} & \textbf{25.59} & \textbf{31.28} & \textbf{18.53} \\
        \bottomrule
        \label{encoder}
    \end{tabular}
\end{table}

\begin{table}[htbp]
    \centering
    \caption{Performance comparison of different ODE solvers.}
    \small
    \setlength{\tabcolsep}{4pt}
    \label{solvers}
    \renewcommand{\arraystretch}{0.85}
    \begin{tabular}{@{}l cc cc cc@{}} 
        \toprule
        \multirow{2}{*}{ODE Solver} & \multicolumn{2}{c}{England} & \multicolumn{2}{c}{Italy} & \multicolumn{2}{c}{Inference Phase} \\
        \cmidrule(lr){2-3} \cmidrule(lr){4-5} \cmidrule(lr){6-7}
        &RMSE & MAE & RMSE & MAE & Time (s) & Speedup \\
        \midrule
        Euler   & 8.09 & 5.69 & 31.28 & 18.54 & 0.52 & 1.00$\times$ \\
        Bosh3   & 8.08 & 5.69 & 31.28 & 18.54 & 0.70 & 0.74$\times$ \\
        Rk4     & 8.08 & 5.67 & 31.28 & 18.53 & 0.71 & 0.73$\times$ \\
        Dopri5  & 8.08 & 5.67 & 31.28 & 18.53 & 0.97 & 0.54$\times$ \\
        \bottomrule
    \end{tabular}
\end{table}
\subsection{Model Stability Analysis}
\paragraph{ODE solver stability.}
To evaluated the computational robustness and runtime differences of Epi$^2$-Net, we implement the model via four distinct numerical solvers (Euler, Bosh3, Rk4, and Dopri5).
Table~\ref{solvers} shows the performance comparison of different ODE solvers, note that the inference time here refers to the runtime on the test set.
We observe that Epi$^2$-Net exhibits consistent performance across various numerical solvers.
Both Rk4 and Dopri5, as high-order Runge-Kutta solvers, demonstrate superior performance compared to lower-order methods. 
We recommend them since their accuracy improvements justify the computational investment for precision-critical forecasting.

\paragraph{Numerical stability.}
Given Epi$^2$-Net's integration of neural ODE module, a temporal encoder, learnable coefficients and a fully-connected layer, its numerical stability must be verified.
Detailed numerical stability analysis and additional stability experiments are presented in Appendix H.

\section{Conclusions}
In this paper, we present Epi$^2$-Net, a physic-inspired deep learning framework to advance epidemic dynamics modeling.
By integrating physical priors into neural architecture, we introduce the neural epidemic transport to build upon Epi$^2$-Net, and reconcile explicit physical prior with neural representations.
Extensive experiment results have demonstrated its superior
performance for epidemic forecasting.

\section{Limitations and Future Works}
Although Epi$^2$-Net integrates case numbers, human mobility, and geographical data, epidemic modeling is a complex problem. Government policies, viral mutation, emergent outbreak, etc., are all key factors for real-world epidemics. 
The current framework struggles to achieve broad integration. Since it cannot fully unlock the potential of data-driven paradigm, combining advanced architectures (Transformer, LLM) with broader multifaceted data is a promising direction. Further discussions are provided in Appendix I.

\bibliography{aaai2026}

\appendix
\newpage

\section{Further Details of Datasets}
\paragraph{Datasets description.}
The COVID-19, first detected in Wuhan, China in December 2019, is caused by the novel severe acute respiratory syndrome coronavirus 2 (SARS-CoV-2), which shares genetic similarities with bat coronaviruses, pangolin coronaviruses, and SARS-CoV.
In this paper, we focus on the COVID-19 epidemic dynamics in selected European countries: England, France, Italy, and Spain.
The dataset comprises three primary components: daily cases, human mobility, and geographical distances.
Basic statistics of datasets are summarized in Table \ref{datasets}.

\paragraph{Daily cases.} Daily confirmed case numbers across part administrative regions of the four study countries are extracted from open-source GitHub repository\footnote{https://dataforgood.fb.com/tools/disease-prevention-maps/}.

\paragraph{Human mobility.} Raw data of human mobility are obtained from anonymized mobile device signals collected through the Facebook App (with user consent and location history enabled), available via a GitHub repository\footnote{https://github.com/geopanag/pandemic\_tgnn}.
The raw dataset contains tri-daily population movement volumes (recorded at midnight, morning, and afternoon) between regions. Following prior studies~\cite{panagopoulos2021transfer}, we aggregated the three measurements into a single inter-regional mobility metric.

\paragraph{Geographical distances.} The inter-regional geographical distances are calculated via centroid coordinates (latitude/longitude) obtained from OpenStreetMap\footnote{https://www.openstreetmap.org}, with spatial distance measurements computed via the Haversine formula.

\begin{table}[ht]
    \small
    \setlength{\tabcolsep}{4pt}
    \centering
    \caption{The statistics information of datasets in epidemic dynamics forecasting for COVID-19.}
    \label{datasets}
    \begin{tabular}{c c c c}
    \toprule
    \textbf{Dataset} & \textbf{Observation Period} & \textbf{\#Regions} & \textbf{Avg.Cases} \\
    \midrule
    England & 2020-03-13 – 2020-05-12 & 129 & 16.7 \\
    France & 2020-03-10 – 2020-05-12 & 81 & 7.5 \\
    Italy & 2020-02-24 – 2020-05-12 & 105 & 25.65 \\
    Spain & 2020-03-12 – 2020-05-12 & 34 & 61 \\
    \bottomrule
    \end{tabular}
\end{table}

\section{Evaluation Metrics}
Let $\mathbf{x}=(x_1,x_2,...,x_n)$ denotes the ground truth, and $\mathbf{\hat{x}} = (\hat{x}_1,\hat{x}_2,...,\hat{x}_n)$ represents the predicted result. 
The evaluation metrics for one region we used in this paper are defined as follows:
\paragraph{Mean Absolute Error (MAE).}
\begin{equation}
    MAE (\mathbf{x},\mathbf{\hat{x}})=\frac{1}{n} \sum_{i=1}^{n} \vert x_i - \hat{x}_i \vert
\end{equation}
\paragraph{Root Mean Square Error (RMSE).}
\begin{equation}
    RMSE(\mathbf{x},\mathbf{\hat{x}})= \sqrt{\frac{1}{n} \sum_{i=1}^{n} ( x_i - \hat{x}_i )^2}
\end{equation}

\section{Implementation Details}
\paragraph{Model implementation.} Epi$^2$-Net is implemented in PyTorch 2.5.1 using NVIDIA GeForce RTX
4090 GPU. 
In the default settings, we use GRU to implement the RNN-based encoder and adopt a non-neural decoder.
We follow~\cite{hettige2024airphynet} and implement the graph Laplacian operators of the Neural ODE module in the form of GCN layers.
For the ODESolver, we use Dopri5, an explicit Runge-Kutta method.
Moreover, models involving neural network modules in the baseline are subjected to five experimental runs, and their average results are reported for evaluation.

\paragraph{Hyperparameter settings.} 
We tune the key hyperparameters in our implementation to achieve optimal model performance. 
We set the batch size to 32, and the initial learning rate to 5e-2. 
We set the GRU encoder hidden dimension to 16, and the observation window length $w$ of encoder to 7.
We used the Adam optimizer with a MultiStepLR learning rate scheduling strategy. 
The model was trained for a maximum of 100 epochs, with an early stopping patience of 10. 
During the encoding process, we employ the reparameterization trick~\cite{ji2022stden,hettige2024airphynet} and generate multiple samples from the latent trajectory to model uncertainty and provide a more robust prediction.
The relative and absolute tolerances (rtol and atol) of ODESolver for Dopri5 are set to 1e-5.
The following list details the search space for each hyperparameter and the final selected optimal values:
\begin{itemize}
    \item Learning rate (lr): \{1e-1, \textbf{5e-2}, 1e-2, 5e-3, 1e-3\}
    \item Lr scheduler gamma: \{0.1, 0.3, \textbf{0.5}, 0.7\}
    \item Lr scheduler milestones: \{[25, 50, 75], \textbf{[25, 35, 45, 55]}, [20, 40, 60, 80]\}
    \item Batch size: \{8, \textbf{16}, 32, 64\}
    \item Hidden dimension of encoder: \{8, \textbf{16}, 32, 64\} 
\end{itemize}

\paragraph{Datasets splits.}
Epidemic forecasting tasks for COVID-19, characterized by rapid outbreak and swift transmission, typically involve spatiotemporal sequences spanning approximately 60 days.
Incompatible with conventional cross-validation settings, we adopt a temporally ordered dataset partitioning strategy, better aligning with real-world epidemic scenarios. Specifically, for our spatiotemporal sequences, we employ a sliding window strategy and partition the data chronologically into training, validation, and testing sets according to a 6:2:2 ratio. This ensures that the validation and test sets are of sufficient size for reliable model selection and performance evaluation, and the entire process strictly adheres to the temporal causality inherent in real-world forecasting tasks. 

\section{Further Details of Baselines}
To evaluate Epi$^2$-Net, we conducted a comparative analysis with 14 methods in epidemic forecasting. The models we benchmark against are as follows:
\begin{itemize}
    \item HA (Historical Average): The epidemic prediction result is derived by averaging the observed cases.
    \item PROPHET~\cite{mahmud2020bangladesh}: A time-series model where the input is the history of entire reported cases for each region, which is widely used in epidemic forecasting.
    \item ARIMA~\cite{kufel2020arima}: An autoregressive moving average model for time-series forecasting, which the input is similar to PROPHET.
    \item LIN\_REG~\cite{kaur2022forecasting}: Given the history of reported cases for each region as input, ordinary least squares linear regression is used to fit the line of cases on the training sets to forecast the future epidemic trend.
    \item RAND\_FOREST~\cite{galasso2022random}: A random forest regression model that produces epidemic forecasting using decision trees, with multiple trees built based on the training sets to best average the final results.
    \item XGBOOST~\cite{fang2022application}: An enhanced version of random forest regression model for epidemic forecasting via gradient boosting.
    \item ITRANSFORMER~\cite{liu2023itransformer}: iTransformer reverses the conventional Transformer's method of processing time series data, redirecting attention from the temporal axis to the feature axis. This framework is then utilized to model the time evolution of case counts for epidemic forecasting.
    \item TIMEMIXER~\cite{wang2024timemixer}: TimeMixer is built upon MLPs, avoiding complex self-attention mechanisms or recurrent structures. It models time series data through multiscale series decomposition and a disentangled mixing strategy. Based on TimeMixer, we model the temporal pattern of case counts for epidemic forecasting.
    \item STGNN~\cite{kapoor2020examining}: This method models human mobility, inter-region connectivity, and epidemic evolution patterns within the epidemic using Spatio-Temporal Graph Neural Networks, to further predict the epidemic.
    \item ATMGNN~\cite{nguyen2023predicting}: A hybrid deep learning model for epidemic forecasting, where multiple resolution GNN are combined with Transformers for modeling the epidemics.
    \item ODE-LSTM~\cite{lechner2020learning}: ODE-LSTM combines RNN structures with Neural ODEs, allowing for the encoding of a continuous-time dynamical flow within the RNN and handling inputs arriving at arbitrary time-lags. By capturing the temporal dependencies of daily new cases, we apply it to epidemic forecasting.
    \item GN-ODE~\cite{kosma2023neural}: The method draws on recent advances in Neural ODEs to design a physics-informed neural network model that approximates the ideal SIR model in networks, thereby modeling contagion dynamics on large complex networks.
    \item EPINNs~\cite{rodriguez2023einns}: Drawing on the work of Physics-Informed Neural Networks (PINNs), EPINNs employs a dual-network structure to achieve epidemic prediction: a physics-constraints (SEIR) network learns latent epidemic dynamics, which is then combined with another neural network module capable of ingesting multiple data sources for collaborative prediction.
    \item EARTH~\cite{wanearth2025}: EARTH integrates Neural ODEs with epidemiological SIR mechanisms to learn continuous regional transmission patterns. It then utilizes a cross-attention mechanism to fuse global and local key information, thereby providing a flexible approach to predicting epidemic spread.
\end{itemize}

\section{Performance Comparison}
We present a additional comparison of Epi$^2$-Net with other baselines on four real-world epidemic datasets.
From Table \ref{results-rmse},
Epi$^2$-Net continues to demonstrate superiority across all four datasets on metrics RMSE. 
It can be concluded that Epi$^2$-Net achieves sota performance in epidemic dynamics modeling, as evidenced by both RMSE and MAE metrics.

\begin{table*}[ht]
    \caption{The overall performance comparison of epidemic dynamics forecasting on metric RMSE. The bold and underlined fonts show the best and the second best result. The improvement(\%) is calculated based on these two results.}
    \label{results-rmse}
    \small
    \setlength{\tabcolsep}{2mm}
    \renewcommand{\arraystretch}{0.9}
    \begin{tabular}{c | c c c | c c c | c c c | c c c }
    \toprule
    \multirow{2}{*}{\textbf{Model}} & 
    \multicolumn{3}{c|}{\textbf{England}} &
    \multicolumn{3}{c|}{\textbf{France}} &
    \multicolumn{3}{c|}{\textbf{Italy}} &
    \multicolumn{3}{c}{\textbf{Spain}} \\ 
    \cmidrule(lr){2-4} \cmidrule(lr){5-7} \cmidrule(lr){8-10} \cmidrule(lr){11-13}
    & 3 days & 5 days & 7 days & 3 days & 5 days & 7 days & 3 days & 5 days & 7 days & 3 days & 5 days & 7 days \\
    \midrule
        HA & 9.68 & 10.21 & 10.64 & 11.49 & 11.61 & 11.81 & 48.97 & 48.06 & 47.62 & 135.95 & 136.29 & 136.19 \\
        PROPHET & 12.67 & 13.58 & 17.73 & 15.79 & 16.60 & 17.12 & 76.96 & 79.83 & 81.81 & 115.44 & 116.89 & 116.58 \\
        ARIMA & 9.05 & 9.88 & 9.82 & 11.87 & 11.36 & 11.63 & 79.13 & 68.08 & 65.76 & 108.87 & 109.12 & 109.92 \\
        \midrule
        LIN\_REG & 11.84 & 12.40 & 12.15 & 10.35 & 10.28 & 10.15 & 52.83 & 52.68 & 53.64 & 120.24 & 124.69 & 122.46 \\
        RAND\_FOREST & 10.87 & 10.05 & 10.36 & 6.05 & 5.89 & 5.52 & 48.14 & 49.96 & 49.46 & 65.13 & 67.11 & 62.10 \\
        XGBOOST & 10.58 & 10.25 & 10.14 & 7.20 & 7.44 & 7.32 & 48.51 & 49.67 & 49.47 & 61.24 & 61.93 & 63.41 \\
        \midrule
        ITRANSFORMER & 8.17 & 8.29 & 8.54 & 5.12 & 5.27 & 5.35 & 33.66 & 33.91 & 33.72 & 45.86 & 47.05 & 43.84 \\
        TIMEMIXER & 8.05 & 8.12 & 8.23 & \underline{4.83} & \underline{4.95} & 5.13 & 33.62 & 33.79 & 33.99 & 43.86 & 46.25 & 45.77 \\
        STGNN & 8.96 & 8.83 & 9.32 & 5.54 & 5.25 & 5.27 & 39.40 & 39.92 & 39.94 & 41.64 & 41.15 & 41.89 \\
        ATMGNN & 8.09 & 8.14 & \underline{8.15} & 5.14 & 5.28 & \underline{5.07} & 38.39 & 38.95 & 39.56 & \underline{41.27} & \underline{40.94} & \underline{40.97} \\
    \midrule
        ODE-LSTM & 10.19 & 10.33 & 10.42 & 6.67 & 6.89 & 7.18 & 48.67 & 48.62 & 49.03 & 50.02 & 50.03 & 50.26 \\
        GN-ODE & 11.56 & 11.84 & 11.92 & 9.84 & 9.77 & 10.03 & 45.23 & 45.82 & 45.29 & 52.23 & 51.18 & 55.52 \\
        EPINNs & 8.82 & 9.31 & 9.42 & 5.56 & 5.52 & 5.50 & 33.23 & 33.45 & 33.52 & 44.92 & 44.68 & 45.08 \\
    EARTH & \underline{7.97} & \underline{8.05} & 8.29 & 5.52 & 5.43 & 5.30 & \underline{32.54} & \underline{32.82} & \underline{32.19} & 42.23 & 42.65 & 42.68 \\
    Epi$^2$-Net & \textbf{7.57} & \textbf{7.78} & \textbf{8.08} & \textbf{4.73} & \textbf{4.77} & \textbf{4.52} & \textbf{31.61} & \textbf{31.00} & \textbf{31.28} & \textbf{40.41} & \textbf{40.77} & \textbf{40.05} \\
    \midrule
    Improvement(\%) & 5.01 & 3.35 & 0.85 & 2.07 & 3.63 & 10.84 & 2.85 & 5.54 & 2.82 & 2.08 & 0.41 & 2.24 \\
    \bottomrule
    \end{tabular}  
\end{table*}

\begin{figure}[ht]
    \centering 
        \includegraphics[width=0.47\textwidth]{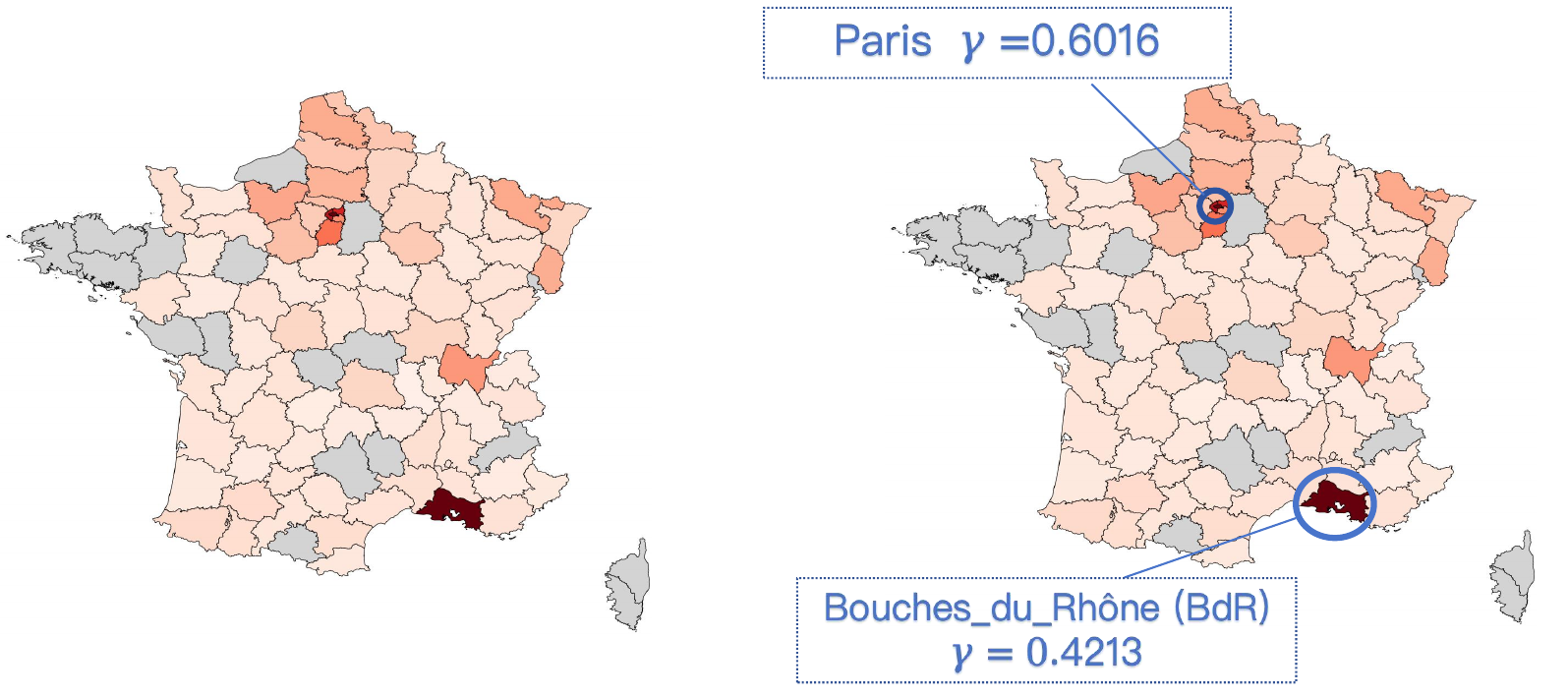} 
    \caption{Visualization of the learned $\gamma$ values for each region from Epi$^2$-Net. The heatmap represents the magnitude of $\gamma$ for each region in France. } 
    \label{gamma} 
\end{figure}
\begin{figure}[ht]
    \centering 
        \includegraphics[width=0.47\textwidth]{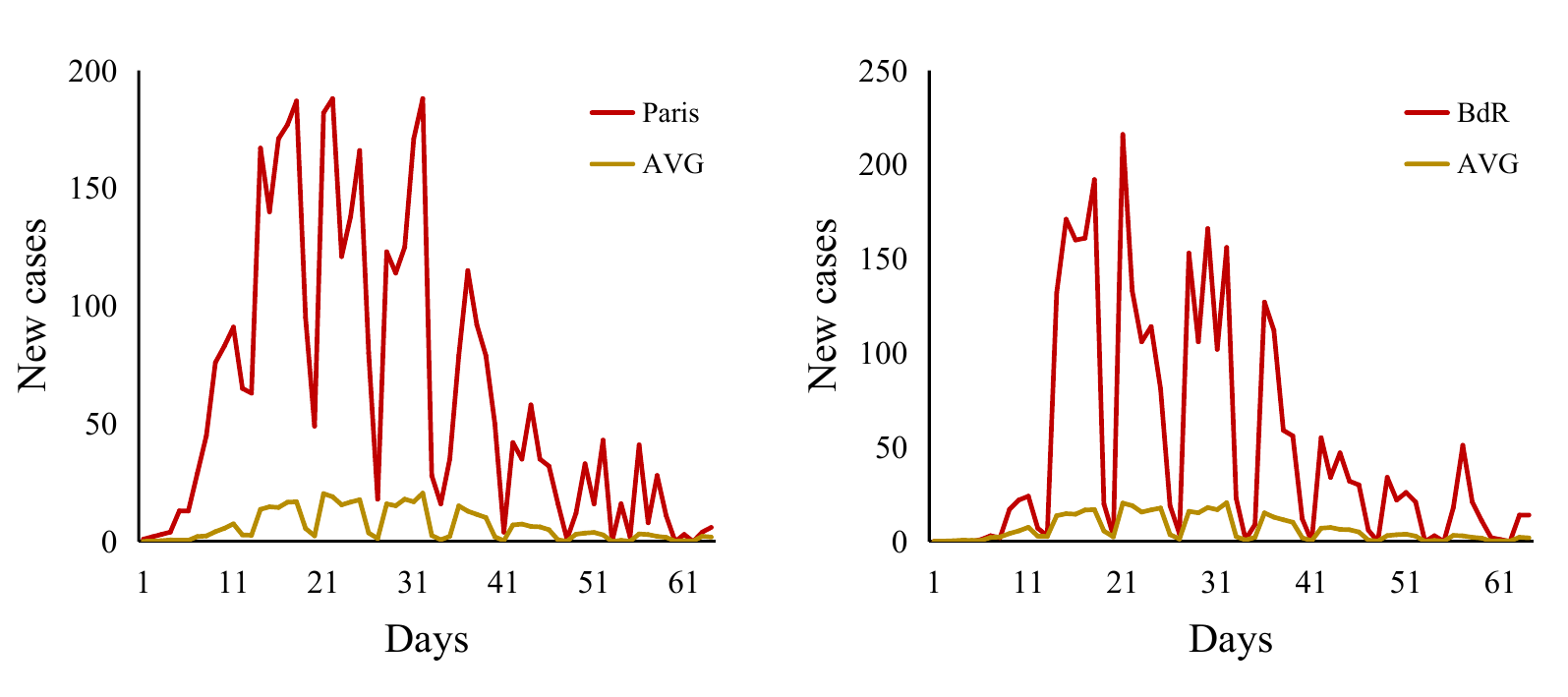} 
    \caption{The infection status of two critical regions: Paris and Bouches-du-Rhône (BdR), and the French national average epidemic situation.} 
    \label{appcs} 
\end{figure}
\begin{figure}[ht]
    \centering 
        \includegraphics[width=0.47\textwidth]{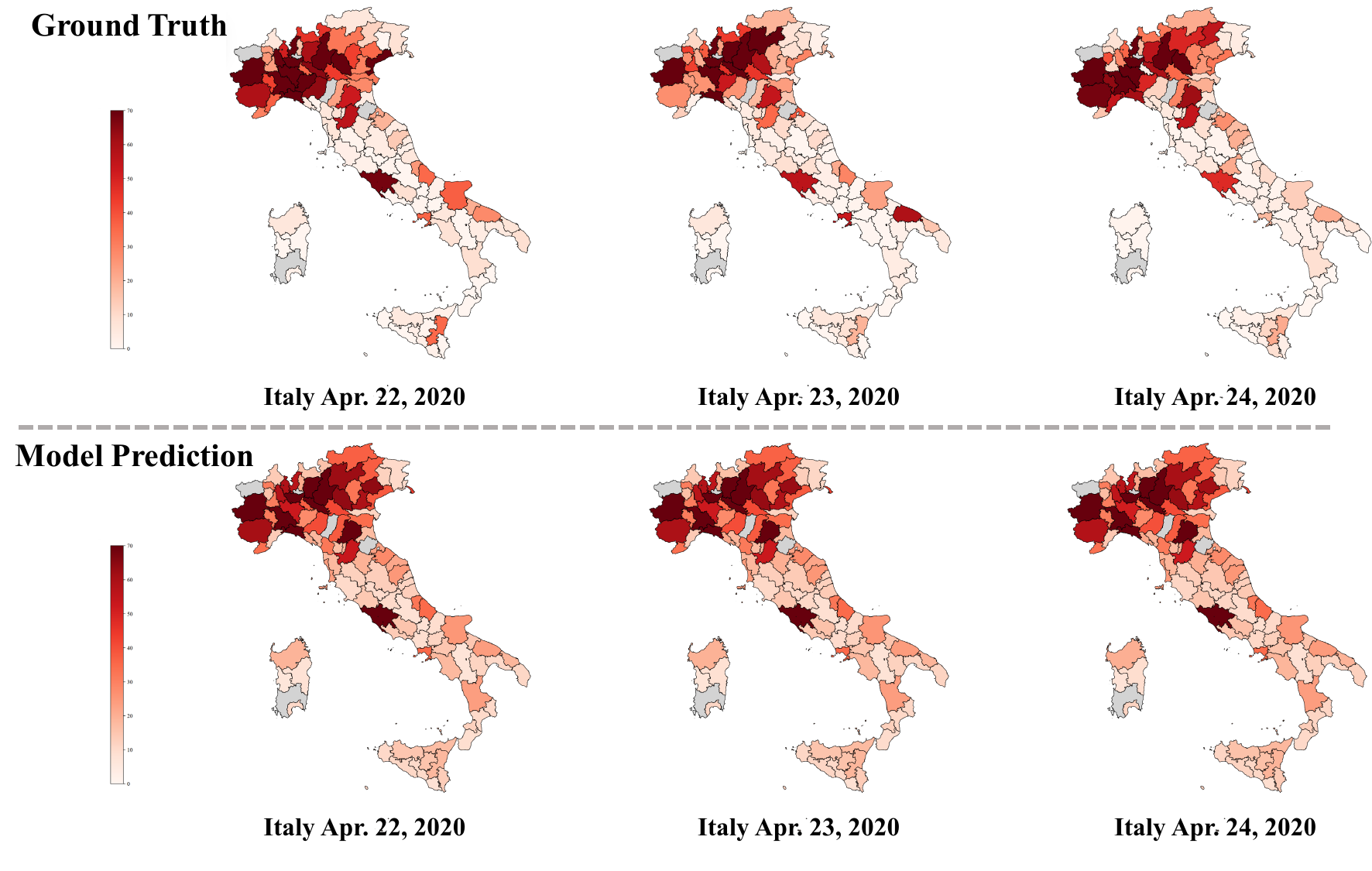} 
    \caption{A case study examines COVID-19 progression in partial regions of Italy. Gray shading denotes areas lacking surveillance records.} 
    \label{case_france} 
\end{figure}
\begin{figure}[ht]
    \centering 
        \includegraphics[width=0.47\textwidth]{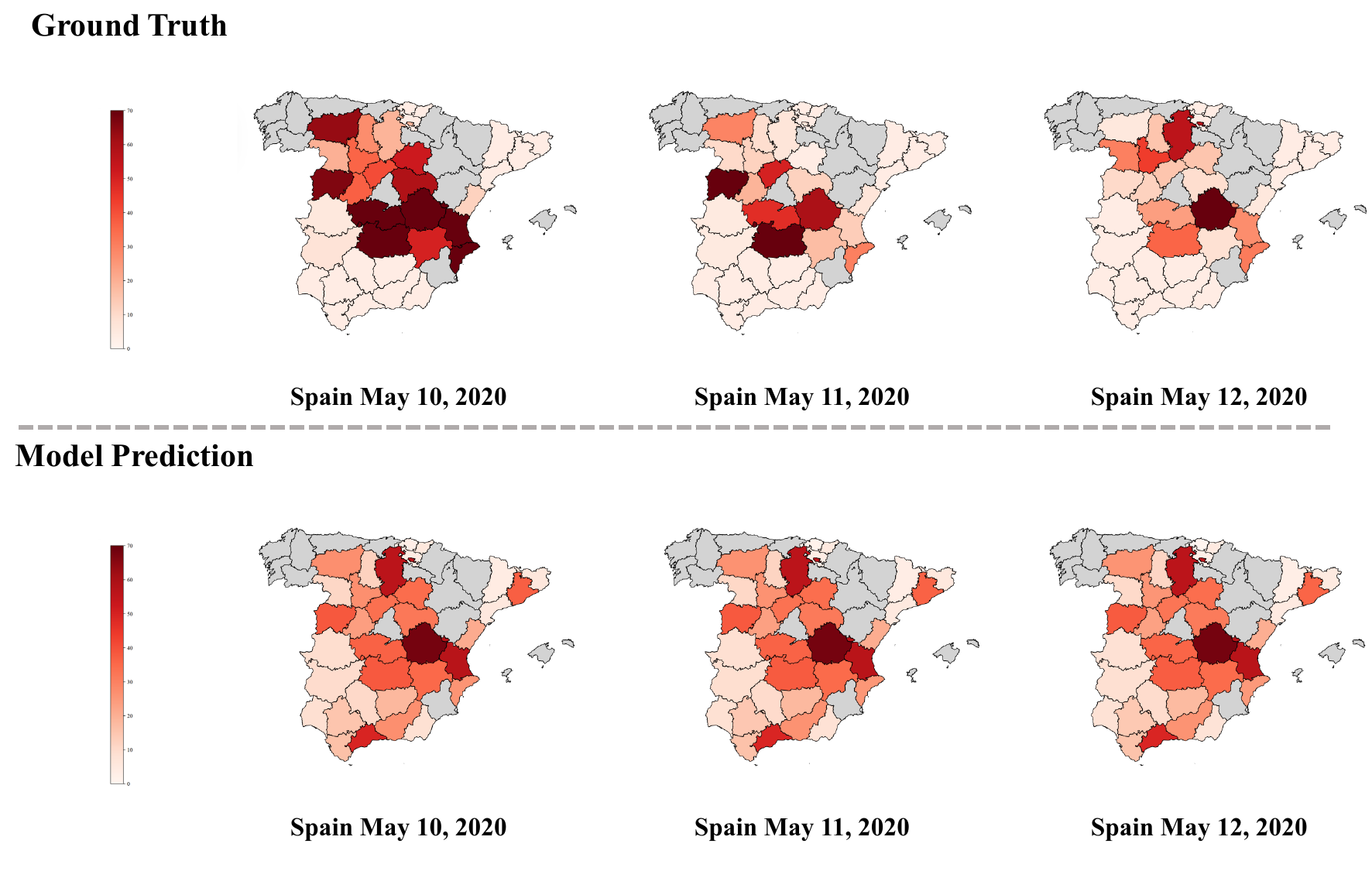} 
    \caption{A case study examines COVID-19 progression in partial regions of Spain. Gray shading denotes areas lacking surveillance records.} 
    \label{case_spain} 
\end{figure}
\section{Case Study}
In addition to the case presented in the main text, we demonstrate Epi$^2$-Net's excellent performance and interpretability for real-world epidemic forecasting via more case analysis in this section.

\paragraph{Model interpretability.}
Given that the learnable reaction term coefficient $\gamma$, within the proposed neural epidemic transport equation, reflects, to a certain extent, a region's infection trend, we visualize the $\gamma$ values learned by Epi$^2$-Net for different regions in France in Figure \ref{gamma}.
Observing the visualization results, it is clear that the $\gamma$ values of two regions are significantly higher than the average level. 
Upon verification, these two regions are identified as Paris and Bouches-du-Rhône (BdR), with their corresponding $\gamma$ being 0.6016 and 0.4213, respectively.
Consequently, we visualized the infection status of these two regions alongside the average infection status across all of France in Figure \ref{appcs}.
It is evident that Paris and Bouches-du-Rhône are severely affected areas (or hotspots) by the COVID-19 epidemic in France.
Considering Paris's central geographical location and BdR's (Bouches-du-Rhône's) port location, the COVID-19 epidemic continued to escalate in these two places. 
Furthermore, the actual measures taken by the French government also corroborate our findings.
As of August 2020, both regions had been designated by the French government as ``highly vulnerable areas'' to the COVID-19 epidemic. 
The preceding discussion highlights the interpretability of the Epi$^2$-Net, suggesting that adopting intervention measures based on its predictions represents a promising approach.

\paragraph{Prediction performance.}
Here, we further elaborate on EpiNet's predictive capabilities through more case studies. 
Beyond the French epidemic discussed in the main text, we have also visualized model's epidemic predictions for Italy, and Spain in Figure \ref{case_france} and Figure \ref{case_spain}.
What we need to emphasize is that in the epidemic
forecasting task for regions in Italy, the predictions made by our model are highly consistent with the actual outcomes.

\begin{table*}[ht]
    \caption{Performance comparison of different temporal encoders on four real-world COVID-19 datasets. The bold and underlined font show the best and the second best result, respectively.
    }
    \small
    \setlength{\tabcolsep}{5pt}
    \centering
    \label{appenc}
    \begin{tabular}{c | c c | c c | c c | c c | c c | c c }
    \toprule
    \multirow{4}{*}{\textbf{Encoder}} & 
    \multicolumn{6}{c|}{\textbf{England}} &
    \multicolumn{6}{c}{\textbf{France}} \\ 
    \cmidrule(r){2-13}
    & \multicolumn{2}{c|}{3 days} & \multicolumn{2}{c|}{5 days} &
    \multicolumn{2}{c|}{7 days} &
    \multicolumn{2}{c|}{3 days} & 
    \multicolumn{2}{c|}{5 days} &
    \multicolumn{2}{c}{7 days} \\
    \cmidrule(r){2-13}
    &RMSE & MAE & RMSE & MAE & RMSE & MAE & RMSE & MAE & RMSE & MAE & RMSE & MAE \\
    \midrule
    LSTM & \underline{7.68}  & \underline{5.46}  & \underline{7.86}  & \underline{5.56}  & \underline{8.12}  & \underline{6.02}  & \underline{4.91}  & \textbf{2.34}  & 4.96  & 2.81  & \underline{4.60}  & \underline{2.66}  \\ 
    TCN & 8.58  & 6.56  & 8.72  & 6.80  & 8.85  & 6.96  & 4.76  & 3.15  & 4.84  & 3.09  & 4.65  & 3.03  \\ 
    Transformer & 9.32  & 7.63  & 9.48  & 7.83  & 9.67  & 7.97  & 5.27  & 3.24  & 5.34  & 3.22  & 5.12  & 3.18  \\
    GRU & \textbf{7.57}  & \textbf{5.35}  & \textbf{7.78}  & \textbf{5.50}  & \textbf{8.08}  & \textbf{5.70}  & \textbf{4.73}  & \underline{2.55}  & \textbf{4.77}  & \textbf{2.62}  & \textbf{4.52}  & \textbf{2.54}  \\
\bottomrule
\end{tabular}

\vspace{0.5em}
\begin{tabular}{c | c c | c c | c c | c c | c c | c c }
    \toprule
    \multirow{4}{*}{\textbf{Encoder}} & 
    \multicolumn{6}{c|}{\textbf{Italy}} &
    \multicolumn{6}{c}{\textbf{Spain}}  \\ 
    \cmidrule(r){2-13}
    & \multicolumn{2}{c|}{3 days} & \multicolumn{2}{c|}{5 days} &
    \multicolumn{2}{c|}{7 days} &
    \multicolumn{2}{c|}{3 days} & 
    \multicolumn{2}{c|}{5 days} &
    \multicolumn{2}{c}{7 days} \\
    \cmidrule(r){2-13}
    &RMSE & MAE & RMSE & MAE & RMSE & MAE & RMSE & MAE & RMSE & MAE & RMSE & MAE \\
    \midrule
    LSTM & \underline{32.18}  & \underline{20.13}  & 34.05  & 21.00  & 34.79  & 21.54  & 47.05  & 37.14  & 47.40  & 36.81  & 47.24  & 36.38  \\ 
    TCN & 33.80  & 20.68  & \underline{32.56}  & \underline{20.72}  & \underline{33.30}  & \underline{21.23}  & 53.84  & 38.53  & 54.70  & 38.89  & 54.74  & 38.61  \\ 
    Transformer & 32.79  & 24.00  & 32.86  & 24.48  & 33.62  & 25.18  & \underline{41.41}  & \underline{33.03}  & \underline{41.77}  & \underline{34.02}  & \underline{41.05}  & \underline{34.13}  \\
    GRU & \textbf{31.61}  & \textbf{18.14}  & \textbf{31.00}  & \textbf{18.21}  & \textbf{31.28}  & \textbf{18.53}  & \textbf{40.41}  & \textbf{25.59}  & \textbf{40.77}  & \textbf{25.19}  & \textbf{40.05}  & \textbf{24.18}  \\
\bottomrule
\end{tabular} 
\end{table*}

\begin{table*}[ht]
    \caption{Performance comparison of different decoders on four real-world COVID-19 datasets. The bold and underlined font show the best and the second best result, respectively.
    }
    \small
    \setlength{\tabcolsep}{5pt}
    \centering
    \label{appdec}
    \begin{tabular}{c | c c | c c | c c | c c | c c | c c }
    \toprule
    \multirow{4}{*}{\textbf{Encoder}} & 
    \multicolumn{6}{c|}{\textbf{England}} &
    \multicolumn{6}{c}{\textbf{France}} \\ 
    \cmidrule(r){2-13}
    & \multicolumn{2}{c|}{3 days} & \multicolumn{2}{c|}{5 days} &
    \multicolumn{2}{c|}{7 days} &
    \multicolumn{2}{c|}{3 days} & 
    \multicolumn{2}{c|}{5 days} &
    \multicolumn{2}{c}{7 days} \\
    \cmidrule(r){2-13}
    &RMSE & MAE & RMSE & MAE & RMSE & MAE & RMSE & MAE & RMSE & MAE & RMSE & MAE \\
    \midrule
    Linear & 7.68  & 5.42  & 7.80  & 5.61  & 8.16  & 5.84  & 4.92  & 2.98  & 4.85  & 2.64  & 4.63  & 2.72  \\ 
    MLP & \textbf{7.07}  & \textbf{5.12}  & \textbf{7.08}  & \textbf{5.13}  & \textbf{7.18}  & \textbf{5.20}  & \underline{4.79}  & \underline{2.96}  & \textbf{4.52}  & \textbf{2.84}  & \textbf{4.14}  & \textbf{2.70}  \\ 
    Non-neural & \underline{7.57}  & \underline{5.35}  & \underline{7.78}  & \underline{5.50}  & \underline{8.08}  & \underline{5.70}  & \textbf{4.73}  & \textbf{2.55}  & \underline{4.77}  & \underline{2.62}  & \underline{4.52}  & \underline{2.54}  \\
\bottomrule
\end{tabular}  

\vspace{0.5em}
\begin{tabular}{c | c c | c c | c c | c c | c c | c c }
    \toprule
    \multirow{4}{*}{\textbf{Encoder}} & 
    \multicolumn{6}{c|}{\textbf{Italy}} &
    \multicolumn{6}{c}{\textbf{Spain}} \\ 
    \cmidrule(r){2-13}
    & \multicolumn{2}{c|}{3 days} & \multicolumn{2}{c|}{5 days} &
    \multicolumn{2}{c|}{7 days} &
    \multicolumn{2}{c|}{3 days} & 
    \multicolumn{2}{c|}{5 days} &
    \multicolumn{2}{c}{7 days} \\
    \cmidrule(r){2-13}
    &RMSE & MAE & RMSE & MAE & RMSE & MAE & RMSE & MAE & RMSE & MAE & RMSE & MAE \\
    \midrule
    Linear & \underline{32.29}  & \underline{18.84}  & \underline{31.40}  & \underline{18.17}  & \underline{31.98}  & \underline{18.72}  & 48.20  & 43.78  & 48.54  & 43.73  & 48.74  & 43.94  \\ 
    MLP & 33.73  & 22.67  & 33.28  & 22.95  & 33.81  & 23.41  & \underline{44.07}  & \underline{38.51}  & \underline{44.56}  & \underline{38.84}  & \underline{44.14}  & \underline{38.63}  \\ 
    Non-neural & \textbf{31.61}  & \textbf{18.14}  & \textbf{31.00}  & \textbf{18.21}  & \textbf{31.28}  & \textbf{18.53}  & \textbf{40.41}  & \textbf{25.59}  & \textbf{40.77}  & \textbf{25.19}  & \textbf{40.05}  & \textbf{24.18} \\
\bottomrule
\end{tabular}  
\end{table*}

\section{Ablation Study}
\paragraph{Impact of temporal encoder.}
Due to space limitations, we present partial ablation experiment results for the temporal encoder in the main body.
Table \ref{appenc} presents the complete ablation experiment results on four real-world COVID-19 datasets for the encoder.
Performance comparison reveals that adopting GRU as the encoder offers advantages across all datasets. Therefore, we employ it for capturing temporal patterns within the Epi$^2$-Net framework.
Furthermore, using LSTM as an encoder can also achieve competitive performance, while TCN as an encoder performs poorly.
Surprisingly, the advanced Transformer architecture performed poorly in our scenario.
We infer that this might be because our observation window length for encoder is not long (default set to 7), so the Transformer's powerful long-range modeling capabilities cannot be well demonstrated.

\paragraph{Impact of decoder.}
In our implementation, we refer to \cite{ji2022stden,hettige2024airphynet} and adopt a parameter-free decoder to generate epidemic dynamics predictions.
To study the efficacy of the decoder, we compare our model
with these variants:
1) Linear: adopt a linear layer as the decoder
2) MLP: adopt a two layer-MLP as the decoder
3) Non-nerual: adopt reshaping and aggregation steps to produce the desired
output format. 
Table \ref{appdec} presents the complete ablation experiment results on four real-world COVID-19 datasets for the decoder.
According to the results, We can observe that using a non-nerual decoder achieved excellent performance in most cases; the MLP decoder performed superiorly on the England and France datasets but poorly on the Italy dataset; the Linear decoder performed slightly worse compared to the MLP and non-nerual decoders.
Therefore, we recommend using a parameter-free operation to implement the decoder in Epi$^2$-Net, due to its excellent and relatively stable performance.

\section{Model Stability}
\begin{figure*}[htbp]
    \centering
    \begin{subfigure}[b]{0.23\textwidth} 
        \centering
        \includegraphics[width=\textwidth]{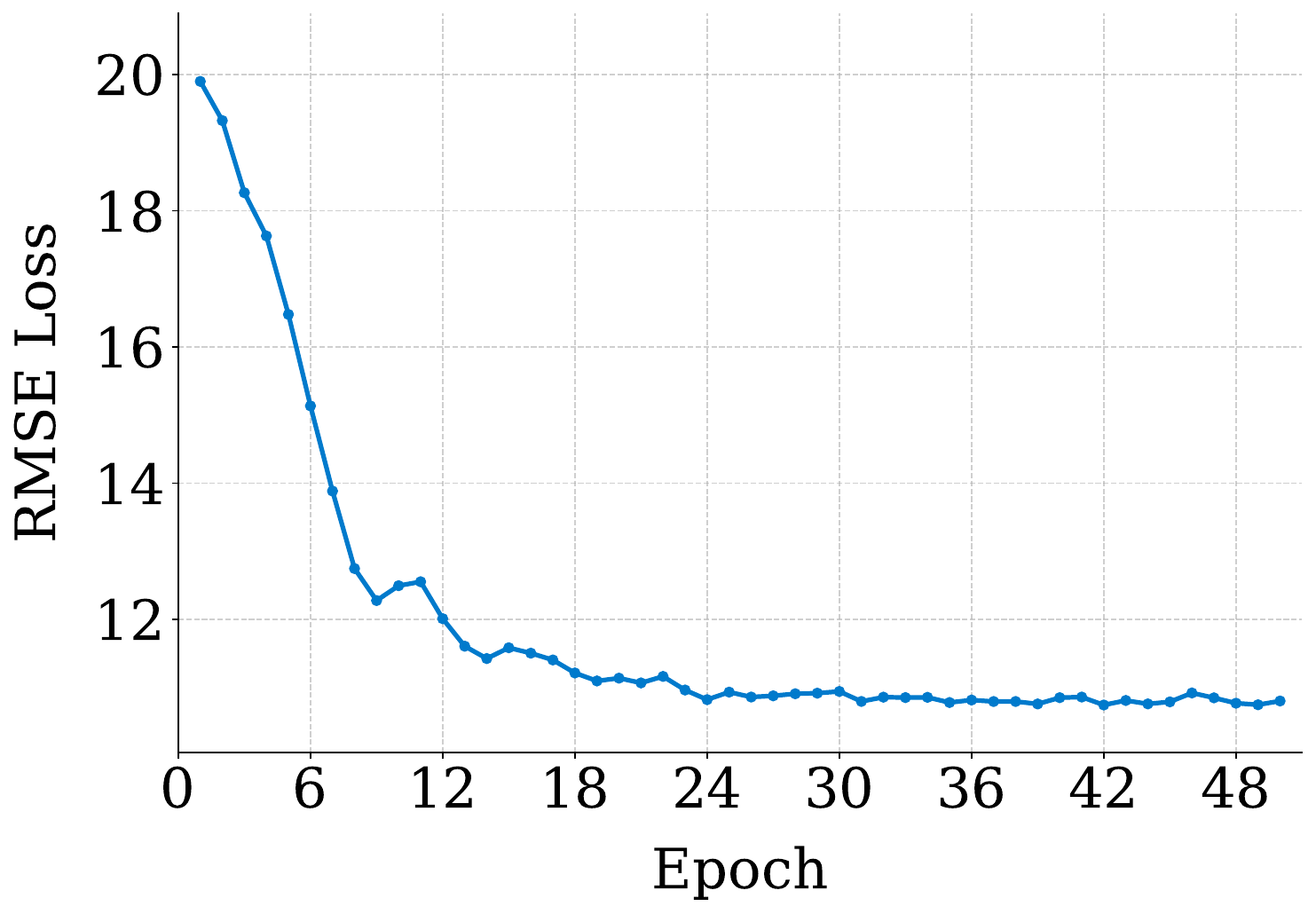} 
        \subcaption[]{England}
    \end{subfigure}
    \begin{subfigure}[b]{0.23\textwidth} 
        \centering
        \includegraphics[width=\textwidth]{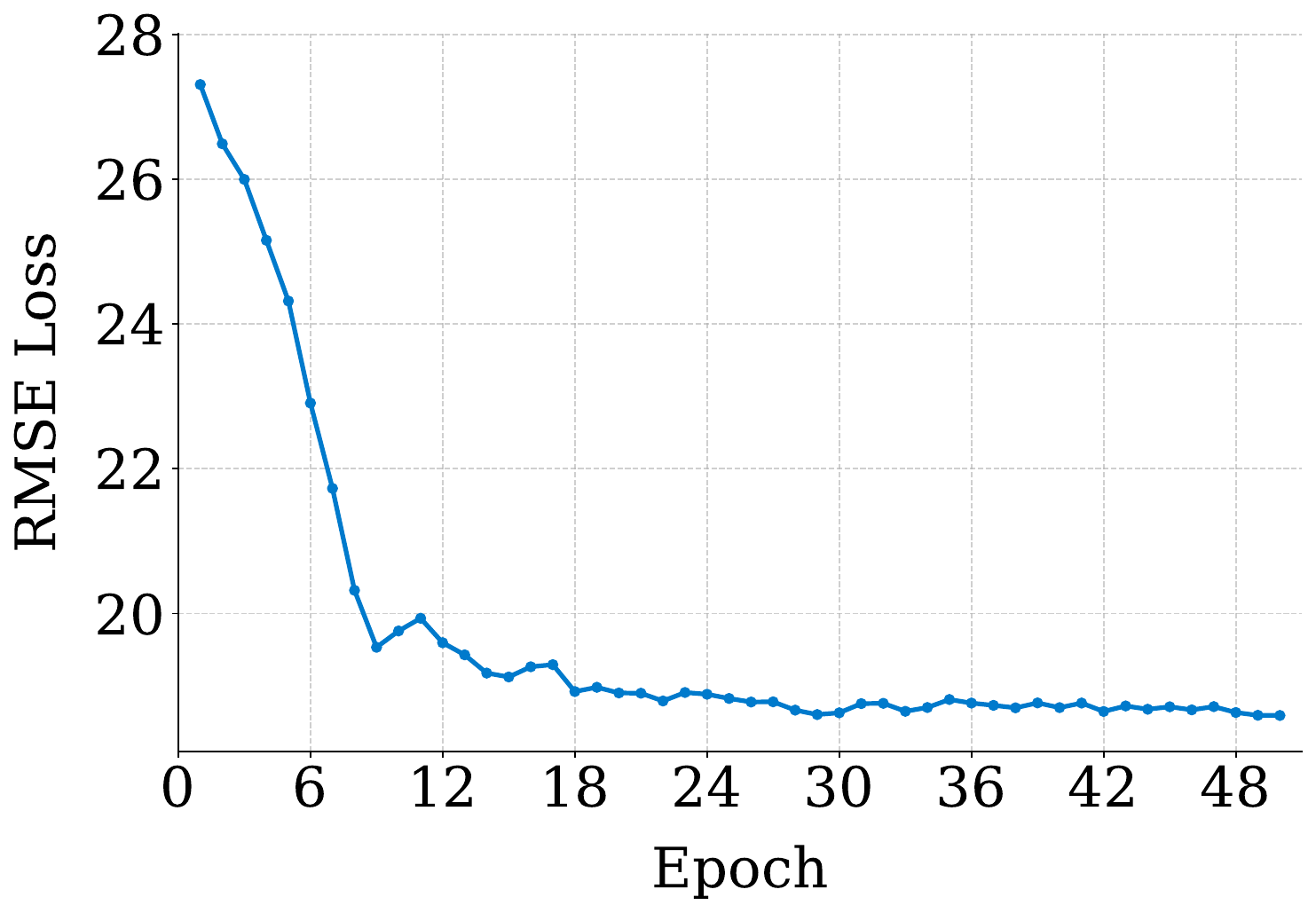} 
        \subcaption{France}
    \end{subfigure}
    \begin{subfigure}[b]{0.23\textwidth}
        \centering
        \includegraphics[width=\textwidth]{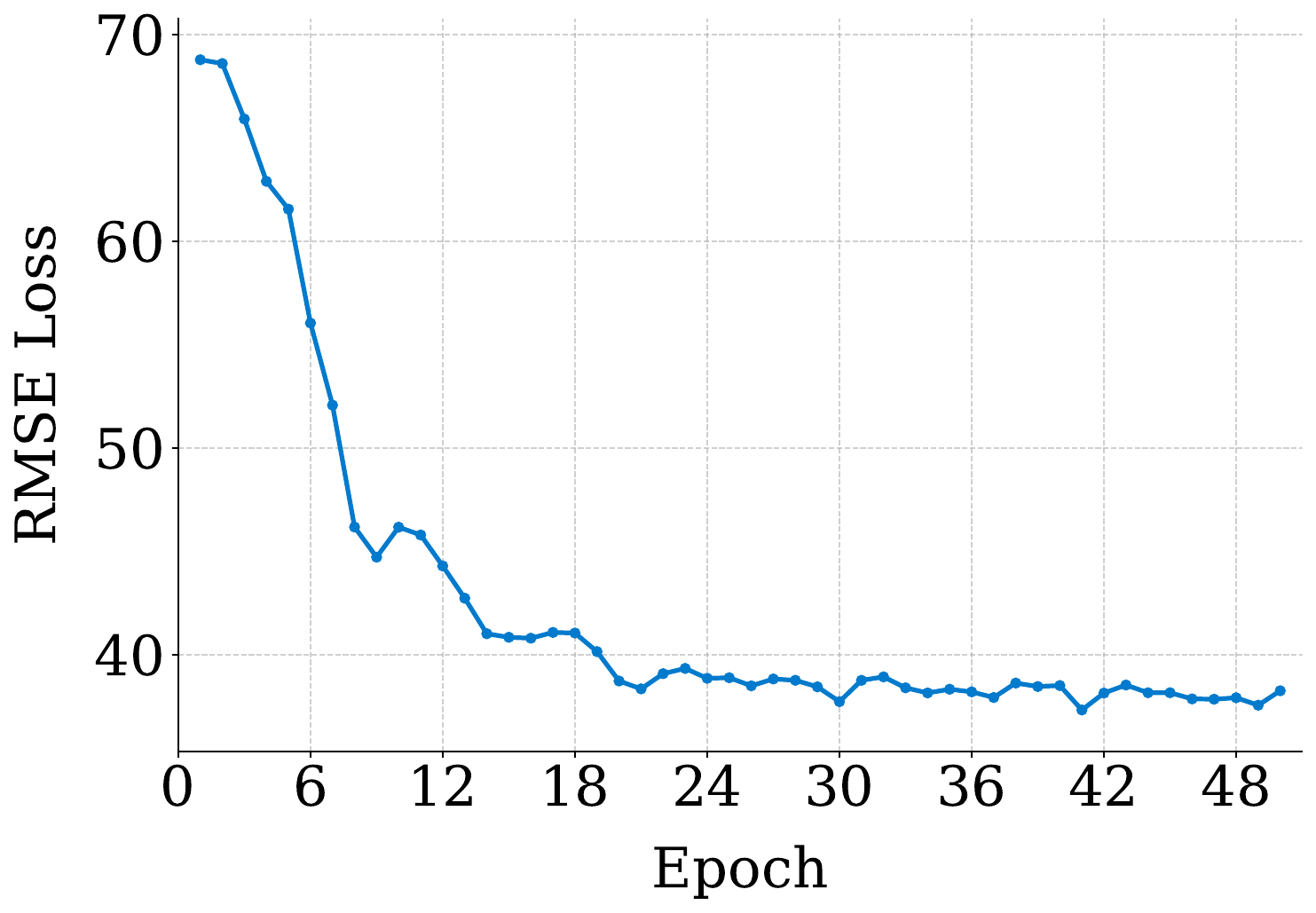}
        \subcaption{Italy}
    \end{subfigure}
    \begin{subfigure}[b]{0.23\textwidth}
        \centering
        \includegraphics[width=\textwidth]{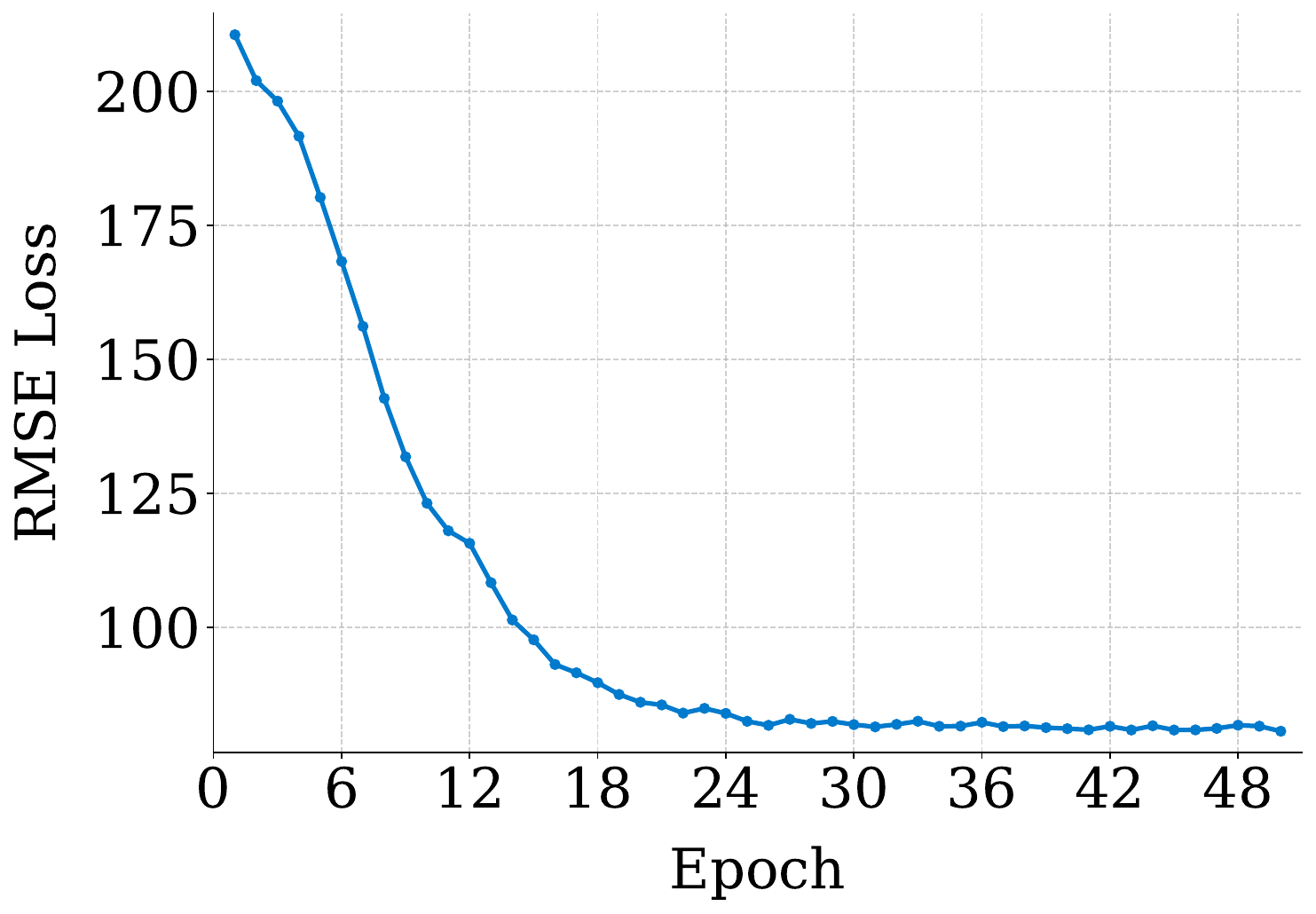}
        \subcaption[]{Spain}
    \end{subfigure}
    \caption{Numerical stability experiments of Epi$^2$-Net on four real-world COVID-19 datasets.} 
    \label{appnum_stb} 
\end{figure*}

\paragraph{Numerical stability.} 
Here, we present the numerical stability experiments of Epi$^2$-Net.
Due to its intricate architecture, which integrates modules including a Neural ODE Module, a temporal encoder, learnable coefficients, and a fully-connected layer, Epi$^2$-Net's numerical stability requires verification.
Figures \ref{appnum_stb} illustrate Epi-Net's training loss curves on four datasets. 
From these, it is evident that Epi-Net's loss rapidly declines and converges during training, thereby showcasing the numerical stability of its model.

\begin{figure*}[htbp]
    \centering
    \begin{subfigure}[b]{0.23\textwidth}
        \centering
        \includegraphics[width=\textwidth]{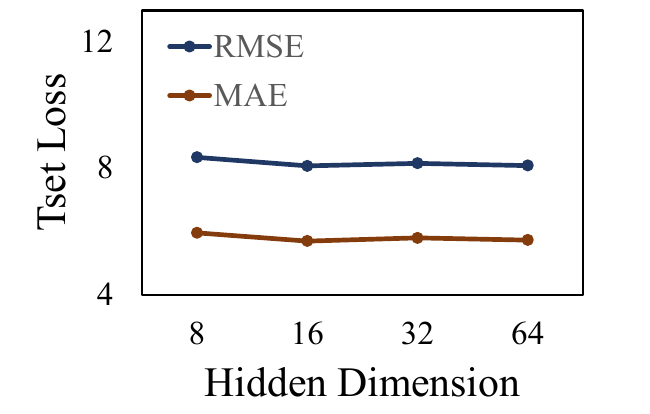}
        \subcaption[]{England}
    \end{subfigure}
    \begin{subfigure}[b]{0.23\textwidth}
        \centering
        \includegraphics[width=\textwidth]{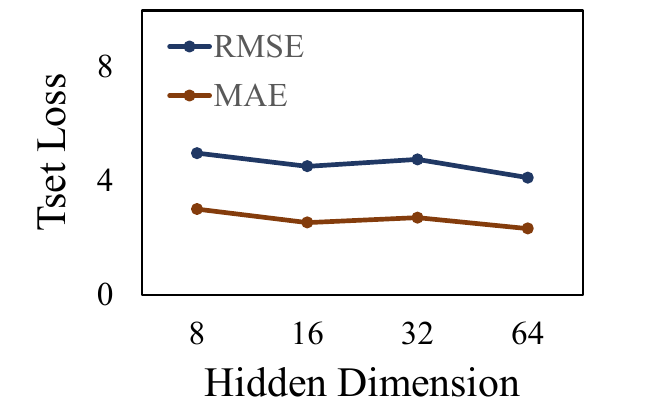}
        \subcaption[]{France}
    \end{subfigure}
    \begin{subfigure}[b]{0.23\textwidth} 
        \centering
        \includegraphics[width=\textwidth]{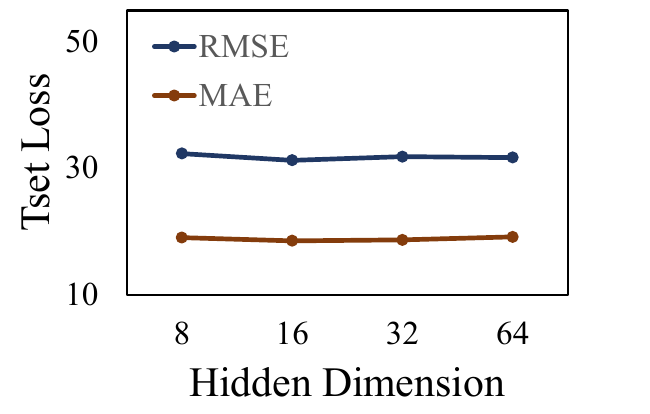} 
        \subcaption[]{Italy}
    \end{subfigure}
    \begin{subfigure}[b]{0.23\textwidth}
        \centering
        \includegraphics[width=\textwidth]{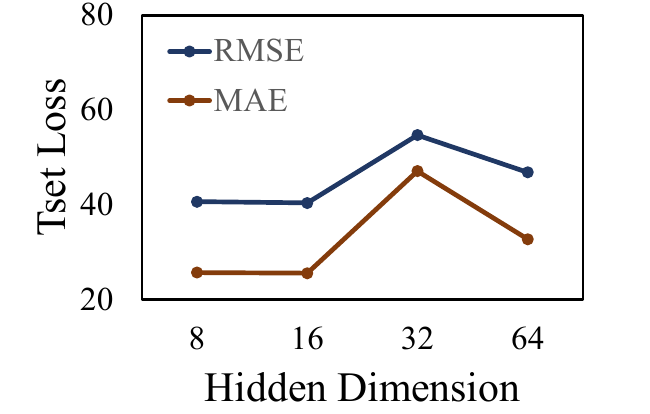}
        \subcaption[]{Spain}
    \end{subfigure}
    
    \caption{Hyperparameter sensitivity analysis of hidden dimension for the GRU encoder in the Epi$^2$-Net.} 
    \label{app_sensi} 
\end{figure*}
\paragraph{Hidden dimension of encoder.}
To assess Epi$^2$-Net's robustness to hyperparameter selection, we performed a sensitivity analysis on the hidden dimension of the GRU encoder.
As shown in Figure~\ref{app_sensi}, the hidden dimension exhibits low sensitivity across the four real-world COVID-19 datasets, consistently achieving good performance.

\section{Discussions}

In the main text, we briefly summarized the limitations of the Epi$^2$-Net; in this section, we will delve deeper into its potential limitations in real-world scenarios.
Furthermore, by discussing the limitations of this paper, we also hope to inspire thought within the epidemic modeling community and contribute some promising research directions to advance AI-enabled public health.

\paragraph{Complexity for real-world epidemic.}
The current framework models real-world epidemic dynamics based on physical transport, novelly integrating diffusion, convection, and reaction processes with the geographical spread of the epidemic, human mobility-driven dissemination, and regional dynamics. It effectively reconciles neural representations with physical prior knowledge.
However, real-world epidemic scenarios involve more complex factors, such as population vaccination rates, public relapse and hospitalization rates, drug resistance, and the probability of viral mutation. 
Therefore, how to incorporate these factors is both a limitation of the framework proposed in this paper and a question worth considering for future research. 

\paragraph{Data resources and model architecture.}
In addition to the various numerical data mentioned above, we also need to consider how to extend the current framework to utilize cross-modal data such as text-based policy announcements, genomic data of new variant strains, and epidemic rumors on social networks to aid epidemic modeling presents a promising direction for future research~\cite{du2024advancing}.
Perhaps utilizing more advanced architectures (e.g., Transformers or Large Language Models) is a feasible solution, but how to bridge the mismatch between physic-inspired models and advanced deep learning models remains a question worth considering~\cite{tian2024air}.
In our framework, adopting advanced architectures like Transformer as module (see in Appendix F) shows a performance decline; fully unleashing the potential of advanced architectures is a promising research direction~\cite{gong2025epillm}.
However, the prerequisite for all of the above is how to obtain diverse and high-quality datasets in the field of public health. 
We call on the community to collectively build pioneering and sophisticated benchmarks to promote the domain development.

\paragraph{Emergent epidemic outbreak.}
While the forecasting results of Epi$^2$-Net are promising, predicting emergent epidemic outbreak remains a significant challenge for epidemic modeling.
Intuitively, the emergent outbreak of an epidemic refers to a rapid increase in the number of infected individuals within a short period, even affecting associated areas. 
Predicting such emergent outbreaks is crucial, as timely public health interventions can prevent hospital bed shortages, medical resource depletion, and even social panic. 
However, the concept for the task of predicting sudden epidemic outbreaks remains vague, urgently requiring a clear definition and the customization of corresponding models.

\paragraph{Domain generalizations.}
The currently proposed Epi$^2$-Net is a domain-specific model for epidemic dynamics modeling. Considering the strong generalization of the diffusion-convection-reaction equation in fields such as atmospheric motion, ocean science, and traffic prediction, it is worth considering modifying the model for specific scenarios to equip it with powerful domain generalization capabilities.

\end{document}